\documentclass{article} 
\usepackage[T1]{fontenc}
\usepackage{iclr2027_conference,times}
\iclrfinalcopy

\usepackage{amsmath,amsfonts,bm}

\def\eqref#1{equation~\ref{#1}}

\def\1{\bm{1}}

\DeclareMathAlphabet{\mathsfit}{\encodingdefault}{\sfdefault}{m}{sl}
\SetMathAlphabet{\mathsfit}{bold}{\encodingdefault}{\sfdefault}{bx}{n}

\usepackage{amsmath,amssymb,amsthm}
\usepackage{graphicx}
\usepackage{xcolor}
\usepackage[most]{tcolorbox}
\usepackage{booktabs}
\usepackage{multirow}
\usepackage{threeparttable}
\usepackage{wrapfig}
\usepackage{subcaption}
\usepackage{algorithm}
\usepackage{algpseudocode}
\usepackage{float}
\usepackage{enumitem}
\usepackage{tabularx}
\usepackage{xspace}
\usepackage{tikz}
\usepackage{iftex}
\ifXeTeX
  \usepackage{xeCJK}
  \IfFontExistsTF{Times New Roman}{\setmainfont{Times New Roman}}{\setmainfont{TeX Gyre Termes}}
  \IfFontExistsTF{Songti SC}{\setCJKmainfont{Songti SC}}{\setCJKmainfont{FandolSong-Regular}}
  \xeCJKDeclareCharClass{CJK}{"2460->"2464}
  \newenvironment{CJK*}[2]{}{}
\else
  \usepackage{CJKutf8}
\fi    
\usepackage{url}
\usepackage{hyperref}   

\AtBeginDocument{%
  \setlength{\abovedisplayskip}{4pt plus 1pt minus 1pt}%
  \setlength{\belowdisplayskip}{4pt plus 1pt minus 1pt}%
  \setlength{\abovedisplayshortskip}{3pt plus 1pt}%
  \setlength{\belowdisplayshortskip}{3pt plus 1pt}%
}

\definecolor{darkmagenta}{rgb}{0.56, 0.0, 1.0}
\hypersetup{colorlinks,linkcolor={blue},citecolor={darkmagenta},urlcolor={blue}}

\definecolor{Cornsilk}{rgb}{1.0, 0.97, 0.86}
\definecolor{lightgray}{rgb}{1.0, 0.827, 0.278}

\newcounter{takeaway}
\newenvironment{takeaway}[2][]{%
  \refstepcounter{takeaway}%
  \begin{tcolorbox}[breakable, colback=Cornsilk, colframe=lightgray,
    title=\textbf{Takeaway \thetakeaway. #2}]%
  \if\relax\detokenize{#1}\relax\else\label{#1}\fi
}{%
  \end{tcolorbox}%
}

\definecolor{accentcolor}{RGB}{116,54,176}
\definecolor{highlightyellow}{RGB}{255,246,205}
\definecolor{highlightframe}{RGB}{232,190,70}

\newcommand{\keysentence}[1]{%
  \begin{center}
    \begin{tcolorbox}[enhanced, width=0.9\linewidth, colback=highlightyellow,
      colframe=highlightframe, boxrule=0.8pt, arc=6pt,
      left=1.2em, right=1.2em, top=0.7em, bottom=0.7em, halign=center]
      #1
    \end{tcolorbox}
  \end{center}
}

\newcommand{\yhead}[1]{%
  \tcbox[on line, arc=1pt, colback=highlightyellow, colframe=highlightframe,
         boxrule=0pt, leftrule=2pt, boxsep=0pt,
         left=3pt, right=3pt, top=1.8pt, bottom=1.8pt]{#1}}

\newcommand{\ptag}[1]{%
  \tcbox[on line, arc=1pt, colback=accentcolor!15!white, colframe=accentcolor,
         boxrule=0pt, leftrule=2pt, boxsep=0pt,
         left=3pt, right=3pt, top=1.8pt, bottom=1.8pt]{#1}}

\newenvironment{highlightbox}{%
  \begin{tcolorbox}[breakable, enhanced, colback=accentcolor!10!white,
    colframe=accentcolor, boxrule=0.8pt, arc=6pt,
    left=1.2em, right=1.2em, top=0.35em, bottom=0.35em]%
}{%
  \end{tcolorbox}%
}

\definecolor{appendixpurple}{RGB}{116,54,176}
\definecolor{appendixlightpurple}{RGB}{248,244,255}

\newcommand{\appentry}[3]{%
  \hyperref[#3]{\textcolor{appendixpurple}{\Large\textbf{#1}}} &
  \hyperref[#3]{\textcolor{appendixpurple}{\textbf{#2}}} &
  \hyperref[#3]{\textcolor{appendixpurple}{\textbf{\pageref*{#3}}}} \\
}

\newcommand{\appsubentry}[3]{%
  &
  \hspace{1.4em}\hyperref[#3]{\textcolor{appendixpurple}{#1\quad #2}} &
  \hyperref[#3]{\textcolor{appendixpurple}{\pageref*{#3}}} \\
}

\newenvironment{defbox}{%
  \begin{tcolorbox}[enhanced, breakable, colback=accentcolor!10!white, colframe=accentcolor,
    boxrule=0.8pt, arc=6pt, left=1.2em, right=1.2em, top=0.7em, bottom=0.7em]
}{%
  \end{tcolorbox}
}

\definecolor{outlinebg}{RGB}{243,243,243}
\definecolor{outlinerule}{RGB}{180,180,180}
\newenvironment{outline}{%
  \begin{CJK*}{UTF8}{gbsn}%
  \par\vspace{0.3em}%
  \begin{tcolorbox}[breakable, enhanced, colback=outlinebg, colframe=outlinerule,
    boxrule=0.5pt, arc=2pt, left=0.9em, right=0.9em, top=0.6em, bottom=0.6em,
    fontupper=\small]%
  \textbf{写作大纲}\par\vspace{0.35em}%
}{%
  \end{tcolorbox}%
  \par\vspace{0.3em}%
  \end{CJK*}%
}

\newtheorem{theorem}{Theorem}

\theoremstyle{definition}
\newtheorem{definition}[theorem]{Definition}

\newtheorem{remark}[theorem]{Remark}

\newcommand{\methodName}{Jev\xspace}
\newcommand{\circlednum}[1]{%
  \tikz[baseline={(char.base)}]{%
    \node[shape=circle, draw=accentcolor, fill=accentcolor!20!white, text=accentcolor,
          font=\footnotesize\bfseries, inner sep=1pt] (char) {#1};}%
}
\newcommand{\cnum}[1]{\circlednum{#1}}

\title{Can Jev be Your Q or Policy in Reinforcement Learning?}

\author{Yi Ma$^1$,
Tianpei Yang$^2$,
Yaodong Yang$^3$,
Weixun Wang$^4$,
Hongyao Tang$^5$
\\[0.5ex]
$^1${Shanxi University},
$^2${Nanjing University}, 
$^3${The Chinese University of Hong Kong},\\
$^4${Independent Researcher},
$^5${Tianjin University}
\\[1ex]
{\small  \texttt{mayi@sxu.edu.cn,tianpei.yang@nju.edu.cn,yydapple@gmail.com,}}\\  {\small \texttt{wxwang@tju.edu.cn,tanghongyao@tju.edu.cn}}
}
\usepackage{placeins}
\usepackage{fvextra,longtable}
\begin{document}

\maketitle

\begin{abstract}
Foundation models supply reinforcement learning (RL) with priors that mitigate its
longstanding weaknesses in sample efficiency and transfer, but their token-by-token
generation makes queries sequential and costly. Jev, a recently released decision model,
generates nothing and returns calibrated, typed answers in a single forward pass. Existing
work studies foundation models in RL either as models to be trained or as generators to be
prompted, and Jev belongs to neither category, having so far served only as a black box in
single domains. How well such a model decides on its own in RL environments, and how it
can improve RL as a component of training, therefore remain unaddressed. To this end, in
this paper we first examine the requirements that the objects of an RL system place on the
answers they consume, and establish that Jev can fulfill all of them except the cardinal
use of a value function. The remaining objects form positions that admit several roles
each. We then construct algorithms that employ Jev at three of these positions, as a
reference policy, an exploration judge, and a replay rater, to improve sample efficiency,
exploration, and learning performance. Across nine MiniGrid tasks and three Atari games,
training with Jev outperforms a standard RL learner, including where the learner makes no
progress alone, while the model itself remains untrained. To our knowledge, we present the
first use of Jev within the RL learning process and establish a frozen decision model as a
usable component of RL training, inviting further exploration of how Jev and other
advanced decision models can improve RL.
\end{abstract}

\section{Introduction}
\label{sec:intro}

Foundation models bring strong generalization and deep priors from large-scale
pre-training \citep{yang2023foundation}, addressing reinforcement learning's longstanding
weaknesses in sample efficiency and transfer \citep{ma2024eureka,rocamonde2024vlmrm}.
These models, however, share a design assumption: they generate text token by token.
Querying such a model is therefore a sequential act, whose cost scales with the length of
the answer and with the number of questions put to it. Jev \citep{typesafe2026docs}, a
recently released decision model, discards that assumption and generates nothing. Given a
text description of a situation together with a set of typed questions, it returns typed
answers in a single forward pass: a choice over a declared set of options, a point on an
ordered rubric, or a probability.

The difference is not only one of cost. Jev's answers are calibrated and confined to
the spaces its questions declare. These properties fall outside the assumptions of
existing work, which
studies foundation models in RL either as models to be trained
\citep{chen2021decision,feng2024nlrl} or as generators to be prompted
\citep{yao2023react,yao2023tree}. Jev belongs to neither category. Where Jev has
been applied, it has served as a black box in single domains such as
crash-narrative coding, radiology, and video quality
\citep{rafe2026jev,huang2026radiology,robitza2026video,ma2026jevstar,wu2026reflex}, and
the open-source work built on it consists largely of zero-shot demonstrations. The
most systematic evaluation to date is confined to text classification
\citep{ibrahim2026evaluating}. What remains unasked is what a decision model of
this kind can be in RL:
\begin{itemize}[leftmargin=2em, itemsep=2pt, topsep=2pt, parsep=0pt]
  \item[\cnum{1}] \emph{How well does Jev decide when its answers are executed directly?}
  \item[\cnum{2}] \emph{How can Jev improve RL as a component of training?}
\end{itemize}

In this paper, we answer these questions in two steps. We first ask how Jev's function
form maps onto the four objects an RL system needs: a policy, a value, a reward, and a
transition model. Each object performs certain operations on its answers, and Jev can
fill it when those operations are defined on one of its three answer forms. Only the
cardinal use of a value function falls outside, since its update performs arithmetic on
the answer. The objects it can fill group into the positions it can occupy, and each
position admits several roles. We build algorithms on three of these positions: (1)
learning from scratch is sample-inefficient, and Jev serves as a reference policy that
guides the learning of the RL policy; (2) uniform exploration is wasteful when rewards
are sparse, and Jev judges when to explore and which experience to encourage; (3) uniform
replay allocates its updates to uninformative transitions, and Jev rates how much each stored
transition is worth learning from. Across the three, we ask how far a frozen decision
model can improve the sample efficiency of a standard RL learner, the experience it
collects, and the performance it reaches.

In our experiments, we use nine MiniGrid tasks \citep{chevalier2023minigrid} and three
Atari games \citep{bellemare2013ale}.
We first characterize what Jev can do as a
decision maker under the inputs our methods supply and compare it with Qwen3-8B
\citep{yang2025qwen3} under a
matched information interface, so that the comparison measures how each model uses the
evidence rather than how much it is given. We then follow the three channels above to test
whether that capability transfers to a learner with its own observations and training
dynamics, each against its counterpart under a fixed interaction budget.

The main contributions of this work are summarized below:
\vspace{-0.15cm}
\begin{itemize}
  \item We formally examine Jev as a candidate for the functions an RL learner needs, and
        propose the four positions it can occupy, together with the roles within them.
  \item We build algorithms for RL with Jev from three perspectives, using it as a
        reference policy, an exploration judge, and a replay rater, aiming to improve
        sample efficiency, exploration, and learning performance.
  \item To our knowledge, this is the first use of Jev in the RL learning process, and we
        demonstrate its efficacy across nine MiniGrid tasks and three Atari games,
        including where the learner cannot progress alone.
\end{itemize}

\section{Background}
\label{sec:background}

\subsection{Related Work}
\label{sec:related}

\vspace{-0.2cm}
\paragraph{Foundation Models for RL} Foundation models are increasingly woven into
reinforcement learning. Early work used them as policies, generating actions
autoregressively from return and history \citep{chen2021decision} or interleaving
reasoning with acting in text environments \citep{yao2023react}. The scope has since widened \citep{cao2024survey}: LLMs design reward
functions by writing and evolving code \citep{ma2024eureka}, supply reward signals from
vision-language feedback \citep{rocamonde2024vlmrm,wang2024rlvlmf}, guide exploration
with world knowledge \citep{bougie2025exploring}, and act as teachers whose guidance is
distilled into a smaller learner \citep{zhou2024llmteacher,klissarov2025maestromotif}.
A smaller body of work asks whether foundation models can supply value or policy priors
directly \citep{yan2025efficient,zhang2024linvit}, though such priors have so far been
demonstrated in text-based decision making rather than in continuous control or
pixel-based benchmarks.

Querying a generative model for these roles is expensive in one of two ways: the reward
and planning lineages consume many \emph{sequential} generations per decision, roughly
eighty per task in \citet{ma2024eureka}, while the policy lineage consumes a single
autoregressive pass whose cost grows with context length. Jev, by contrast, answers a
batch of typed questions in one fixed-cost forward pass. In this paper, we explore to
what degree such a model can play any of these roles in RL.

\vspace{-0.2cm}
\paragraph{Jev and Its Applications} Jev is a general-purpose decision model in essence.
Given a text description of a situation and typed questions, it returns typed answers
in one forward pass, generating no text. TypeSafe AI describes its training
procedure, RLCD, as an alternative to RLHF and RLVR \citep{typesafe2026jev}. Applications of Jev have followed quickly but narrowly. It has been used as a black
box within a single domain at a time: coding crash narratives \citep{rafe2026jev},
scoring radiology reports \citep{huang2026radiology} and video quality
\citep{robitza2026video}, selecting actions in a real-time strategy game
\citep{ma2026jevstar}, and gating a stronger model's decisions \citep{wu2026reflex}.
The surrounding open-source ecosystem consists largely of zero-shot demonstrations.
The most systematic evaluation of a decision model to date spans eighteen
text-classification tasks against nineteen LLM baselines
\citep{ibrahim2026evaluating}.

None of this work measures Jev on a sequential decision, and none places it inside a
learning loop. Its \emph{capability boundary} in RL is therefore unmapped, and its
\emph{potential as a component of training} untested. This paper addresses both: we
first develop a framework that determines which RL roles a closed decision model can
occupy, and then test its predictions on classic control.

\subsection{Preliminaries and Notation}
\label{sec:prelim}

\vspace{-0.2cm}
\paragraph{Reinforcement Learning} We consider an RL agent interacting with a Markov decision process
$M=(S,A,P,r,\gamma)$ with state space $S$, action space $A$, transition kernel
$P(s'\mid s,a)$, reward function $r$, and discount factor $\gamma\in[0,1)$. The
agent's policy $\pi:S\to\Delta(A)$ induces the action-value function
$Q^{\pi}(s,a)=\mathbb{E}_{\pi}\big[\sum_{t\ge 0}\gamma^{t}r_{t}\mid s_{0}=s,\,a_{0}=a\big]$
and the value function $V^{\pi}(s)=\mathbb{E}_{a\sim\pi(s)}[Q^{\pi}(s,a)]$. The learning
objective is to find a policy maximizing
$J(\pi)=\mathbb{E}_{s_{0}\sim\rho}[V^{\pi}(s_{0})]$, where $\rho$ is the initial state
distribution; we write $\pi^{\star}$ and $Q^{\star}$ for the optimal policy and its
action-value
function \citep{sutton2018reinforcement}.

\vspace{-0.2cm}
\paragraph{The Jev Model} Jev is a closed decision model \citep{typesafe2026docs}. Let
$\mathcal{U}$ denote the space of typed questions and $\mathcal{X}$ the space of texts
describing a \emph{situation}\footnote{The interface names this field \texttt{state}; we
avoid that term to keep it distinct from the MDP state.}, the context against which
a question is posed. A question declares its own answer space
$\mathcal{Y}_{u}$. We write $\mathcal{X}_{Y}\subseteq\mathcal{X}$ for the situations that
describe an element of $Y$, so that a question about a state takes a situation in
$\mathcal{X}_{S}$. A \texttt{choice} question names a finite set $C$ of options and
answers in $\Delta(C)$; a \texttt{score} question names an ordered rubric and answers
with a point on it; a \texttt{noul} question asserts a proposition and answers with a
probability in $[0,1]$. A request pairs a question $u\in\mathcal{U}$ with a situation
$x\in\mathcal{X}$ and returns an answer in $\mathcal{Y}_{u}$.

Note that the question $u$ selects the function computed, as the same situation posed
under different questions yields different functions. We write
\begin{equation}
f_{u}:\mathcal{X}\to\mathcal{Y}_{u},\qquad f_{u}(x)=\mathrm{Jev}(u,x),
\label{eq:jev}
\end{equation}
and work with $f_{u}$ throughout. Since the model is closed, the role it plays is
determined entirely by which function is selected: Jev can act as a policy when the
question asks which action to take, and as a Q function when the question asks how good
an action is.

In fact, a request may carry several questions at the same time, and they are answered
independently of one another. We formalize with a single question per call, so that each
role corresponds to one function. The remaining interface details, such as option and rubric limits, context
budget, latency, and cost, are deferred to Appendix~\ref{app:jev}.

\section{Jev as a Decision Maker: Its Roles in RL}
\label{sec:roles}



In this section, we first examine what an RL system requires of a function approximator,
and which of those requirements Jev's answer forms can meet. We then group the objects
that survive into the positions Jev can occupy.

\subsection{What It Takes to Fulfill a Role in RL}
\label{sec:criteria}

An RL system asks four things of a function approximator: to choose actions
($\pi:S\to\Delta(A)$), to value state-action pairs ($Q:S\times A\to\mathbb{R}$), to reward
states ($R:S\to\mathbb{R}$), and to predict next states
($P:S\times A\to\Delta(S)$). All four take a state-like input and return a number or a
distribution, and each of Jev's three answer forms (i.e., \texttt{choice}, \texttt{score},
and \texttt{noul}) can be fitted to these shapes, since a point on a rubric is a number and
a choice over options is a distribution. Therefore,
what separates the roles is how the functions they represent, with specific descriptions by
the question $u$ work in RL, and a function fills a role when the operations that role
applies to its answers are defined.

Reward is the least demanding of the four. It enters the objective as a sum of terms, so it needs one scale
shared across states, and a \texttt{score} question on a fixed rubric supplies it. A
policy needs a distribution over the action set, and a transition model one over the
state space; a \texttt{choice} question supplies both whenever those sets are finite. The
value function is the most demanding. Its target $r+\gamma\max_{a'}Q(s',a')$ adds a
reward to a value and discounts the sum, and the update fits the prediction to the result,
so the answer has to be a number on a cardinal scale, where intervals are comparable, and
a unit is fixed. A \texttt{score} answer is not, because it is a point on an ordered
rubric, which fixes an order but not the distance between levels. A value can therefore be
ranked but not added, multiplied, or averaged. The ordinal use of a value survives; the
cardinal use does not. Table~\ref{tab:roles} shows the requirements and the verdicts.

\begin{wraptable}{r}{0.52\linewidth}
  \centering
  \vspace{-0.6cm}
  \caption{The answer form required of each object and the verdict.}
  \label{tab:roles}
  \vspace{-0.3cm}
  \small
  \setlength{\tabcolsep}{4pt}
  \begin{tabular}{lll}
    \toprule
    \textbf{Object} & \textbf{Answer Form} & \textbf{Verdict} \\
    \midrule
    $R$           & one scale over states & usable \\
    $\pi$         & distribution over $A$ & usable, $A$ finite \\
    $P$           & distribution over $S$ & usable, $S$ finite \\
    \midrule
    $Q$, cardinal & arithmetic            & \textbf{excluded} \\
    $Q$, ordinal  & an order              & usable as a ranker \\
    \bottomrule
  \end{tabular}
  \vspace{-0.7cm}
\end{wraptable}

Among the four objects, only the cardinal use of a value function is excluded. The others
are admitted subject to conditions the question itself states, namely a declared finite
action set for a policy, a declared finite state space for a transition model, and one
shared rubric for a reward.

\subsection{The Roles Jev Can Play}
\label{sec:roles-list}

Each of the four objects in the previous subsection corresponds to a family of
questions, and hence to a function $f_{u}:\mathcal{X}_{Y}\to\mathcal{Y}_{u}$ of the
situation. We merge roles whose functions coincide in both domain and answer space into a
single \emph{position}. The four positions are as follows.

\begin{highlightbox}
\begin{itemize}[leftmargin=*, itemsep=2pt, parsep=0pt, topsep=0pt, partopsep=0pt]
  \item \textbf{Jev as Policy.} $f_{u_\pi}:\mathcal{X}_S\to\Delta(C)$, where $C$ is a
    finite set of choices named by the question. When $C$ is the action set, Jev fills it
    as an \emph{execution policy} that acts online, or as a \emph{teacher} or \emph{expert}
    that labels states offline.
    The position is available whenever the choice set can be declared.
  \item \textbf{Jev as Score Rater.} $f_{u}:\mathcal{X}_S\to\mathcal{Y}_{u}$, a
    \texttt{score} question that places a state, or a candidate named by the question, on
    an ordered rubric. The same answer can be read in two ways. Read through its order alone,
    it yields a \emph{Q rank} over actions or states that can be used for learning. Read as
    a scale shared by every query, it is a score function that can supply \emph{reward},
    \emph{shaping}, \emph{exploration bonuses}, or \emph{credit assignment}.
  \item \textbf{Jev as Meta Regulator.} $f_{u}:\mathcal{X}\to[0,1]$, a \texttt{noul} question that
    asserts a proposition and returns its probability. The answer does not enter the MDP as
    an action, a reward, or a value; it decides whether something happens, such as
    exploring at the current state, consulting a stronger model, or handing control to a
    human. The proposition may concern the learning process itself, so the situation is not
    always one the environment produced.
  \item \textbf{Jev as Transition.} $f_{u_P}:\mathcal{X}_{S\times A}\to\Delta(S)$, a
    \texttt{choice} question over a declared finite state set. Jev fills it as a
    \emph{transition model} (or \emph{world model}), available for finite $S$ and for short
    rollouts only, since each answer becomes the next question's input and its errors
    accumulate.
\end{itemize}
\end{highlightbox}

Each of these positions carries a different signal into the learning process, and the next
section builds on the first three. The transition position is set aside for now, since it
constrains the state space rather than the signal the learner receives.


\section{Jev for RL: Algorithms}
\label{sec:method}

\begin{figure}
    \centering
    \includegraphics[width=\linewidth]{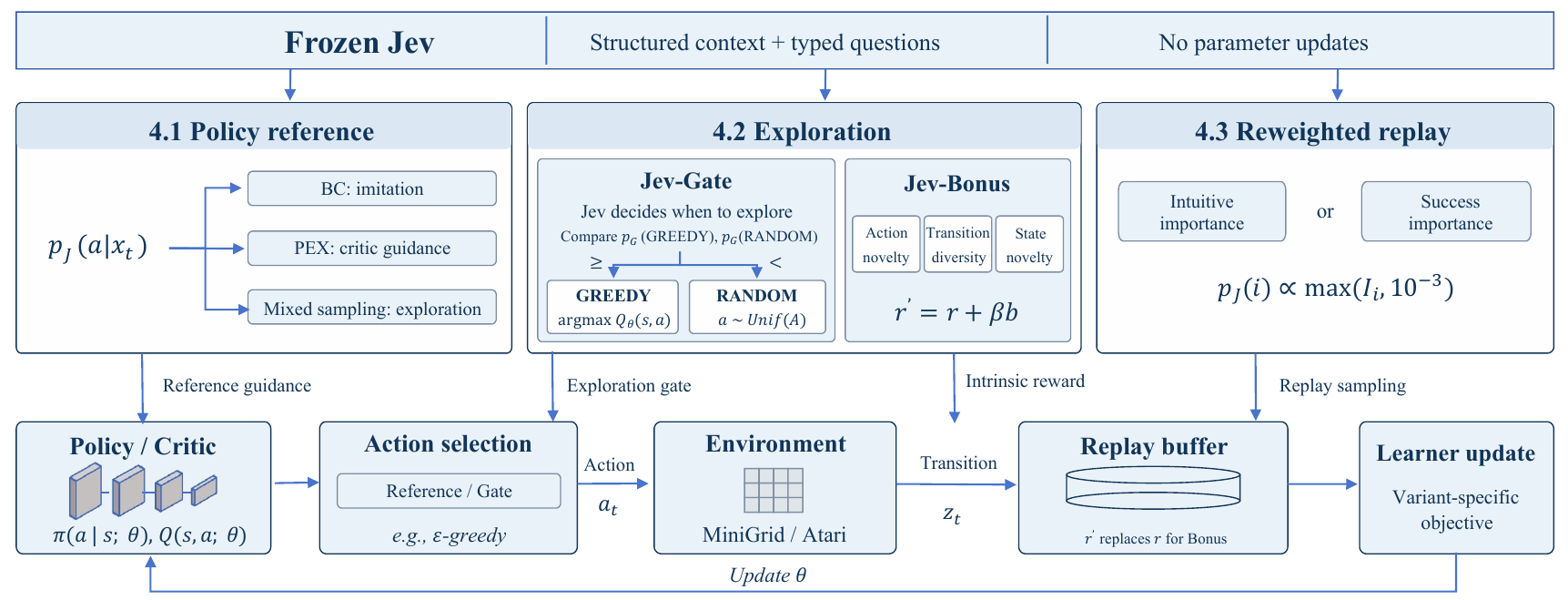}
    \caption{The framework structure of Jev for RL.}
    \label{fig:framework}
\end{figure}

Since Jev can function at the four positions and manifest as different roles as presented
above, it has the potential to enhance RL in many ways. In this section, we propose concrete algorithms
that put it in these roles along three lines (Figure \ref{fig:framework}): (1) Jev serves as a policy reference that
guides the RL policy learning in Sec.~\ref{sec:jev-reference}; (2) Jev shapes
exploration through a gate and novelty bonuses in Sec.~\ref{sec:jev-exploration};
(3) Jev sets the replay priorities of stored experiences for better learning in
Sec.~\ref{sec:jev-sampling}. We
defer the experimental evaluation of these algorithms
to Sec.~\ref{sec:experiments}, and further discussion of RL with Jev to
Sec.~\ref{sec:conclusion}.

\subsection{RL with Jev Reference}
\label{sec:jev-reference}

Classic RL learns from scratch, which is widely known to be sample-inefficient. Introducing
an extrinsic prior reference is one major way to address this point. Imitation learning
trains the student to reproduce a reference's actions
\citep{pomerleau1991,ross2011dagger,hussein2017imitation}; policy guidance keeps the
reference available and lets the student decide when to follow it \citep{zhang2023pex}. Here, we take
the Jev model as the reference, instantiating the two algorithms as Jev-BC and Jev-PEX, which use
the \ptag{Jev as Policy} position of Sec.~\ref{sec:roles-list} with the legal actions declared as the
option set of a \texttt{choice} question.

Let $s_t$ be the agent's observation and $x_t$ the structured input available to Jev. The
reference question asks which legal action is optimal for the current scene and task, returning
$p_J(a\mid x_t)=f_{u_{\mathrm{act}}}(x_t)_a$ for $f_{u_{\mathrm{act}}}:\mathcal{X}_S\to
\Delta(A)$, with $a_J=\arg\max_a p_J(a\mid x_t)$.

\vspace{-0.3cm}
\paragraph{\yhead{Jev-BC.}}
The most direct use of the answer is to reproduce it, storing the reference label with each
collected observation and minimizing
$\mathcal L_{\text{Jev-BC}}(\theta)=
-\mathbb E_{(s_t,a_J)\sim\mathcal D}\log\pi_\theta(a_J\mid s_t)$.
The objective is imitation alone, so the measurement is how much reference behaviour the
student can capture.

\vspace{-0.3cm}
\paragraph{\yhead{Jev-PEX.}}
Jev-PEX instead lets the learner assess that choice, adapting policy expansion
\citep{zhang2023pex} with the reference in place of its fixed offline policy. The minimum of
two critics weighs the reference against a student proposal,
\begin{equation}
 P(c\mid s_t,x_t)=
 \frac{\exp(\eta Q_{\min}(s_t,a_c))}
 {\sum_{c'\in\{J,\pi\}}\exp(\eta Q_{\min}(s_t,a_{c'}))},
 \qquad c\in\{J,\pi\},\quad\eta=10,
 \label{eq:jev-pex}
\end{equation}
and each actor update selects a composite action from Eq.~\ref{eq:jev-pex} and the stored
label. The critic, not the reference, decides utility on the task's return scale.

\vspace{-0.3cm}
\paragraph{\yhead{Jev-Mixed Sampling.}}
A third use bypasses actor training. We keep DQN's action rule and its schedule $\epsilon_t$,
but the action drawn with probability $\epsilon_t$ comes from the reference rather than from a
uniform draw, being either its preference $a_J$ (Jev action) or a sample $a\sim p_J(\cdot\mid x_t)$ (Jev probability); the
actor then learns from the environment's reward alone.

The reference is never updated in any of the three uses, and none of them requires an offline
dataset or a pretrained student. Its influence reaches the learner only through the queries
made at the states the learner visits.

\subsection{RL with Jev Exploration}
\label{sec:jev-exploration}




Learning from scratch in sparse reward settings is extremely difficult, and random exploration does not provide informative experience. Existing exploration methods introduce intrinsic bonuses based on visitation counts \citep{bellemare2016unifying} or prediction errors \citep{pathak2017curiosity,burda2019rnd} to encourage potentially valuable exploration. Here, we take the Jev model as an external exploration judge, using the task description and interaction history to inform when to explore and what experience to encourage. Specifically, we instantiate two uses: \emph{Jev-Gate} and \emph{Jev-Bonus}, which use the \ptag{Jev as Meta Regulator} and \ptag{Jev as Score Rater} positions of Sec.~\ref{sec:roles-list}, respectively.


\vspace{-0.3cm}
\paragraph{\yhead{Jev-Gate.}}
The most direct use of the judgment is to decide whether to follow the policy's greedy action or explore. The gate question receives the current scene, the greedy action and its Q-values, the update count, and an experience history $H_t$. Writing these inputs as $x_t^G$, the question returns
$p_G(\cdot)=f_{u_{\mathrm{gate}}}(x_t^G)$,
with ${\mathrm{GREEDY},\mathrm{RANDOM}}$ declared as the option set of a \texttt{choice} question. We take $\arg\max_a Q_\theta(s_t,a)$ when
$p_G(\mathrm{GREEDY})\geq p_G(\mathrm{RANDOM})$ and $a\sim\mathrm{Unif}(A)$ otherwise. Jev-Gate replaces $\epsilon$-greedy and decides when to explore, while the exploratory action remains uniformly sampled.

\vspace{-0.3cm}
\paragraph{\yhead{Jev-Bonus.}}
A second use of Jev for exploration is to assign intrinsic bonuses to experience. The novelty question asks how unfamiliar the experience is, using a history $H_t$ maintained for that query type, with the query's own past inputs and cumulative visitation counts. Its answer is a \texttt{choice} over
$\mathcal L=\{$zero, low, medium, high, very\_high, unknown$\}$ with values $v=(0,.25,.5,.75,1,0)$, we convert the returned distribution into
$R_J(z;u)=\sum_{\ell\in\mathcal L}v_\ell f_u(z)_\ell\in[0,1]$, where $u$ asks about evidence $z$. We instantiate three variants according to what the question evaluates:
\begin{equation}\begin{aligned}
 b^{\mathrm{act}}_t&=R_J((x_t,a_t,H_t);u_{\mathrm{act\mbox{-}nov}}),\\
 b^{\mathrm{trans}}_t&=R_J((x_t,a_t,x_{t+1},r_t,H_t);u_{\mathrm{div}}),\\
 b^{\mathrm{state}}_t&=R_J((x_{t+1},H_t);u_{\mathrm{state\mbox{-}nov}}).
 \label{eq:jev-novelty}
\end{aligned}\end{equation}

The action bonus judges the novelty of the executed action. The transition bonus instead judges the realized transition, with the observed reward as context. The state bonus judges the reached state novelty given its visitation history. Each variant adds its bonus to the environment reward as $r'_t=r_t+\beta b_t$. We store the bonus with the transition and reuse it during replay. Only the executed action's rating enters training; the bonuses do not select among candidate actions. Tables~\ref{tab:exploration-prompts} and~\ref{tab:query-memory}  specify the instructions and memory used by each question.

These bonuses are auxiliary training objectives, not potential-based rewards with a policy-invariance guarantee, so their empirical value depends on both the content and the scale of the induced bonus.

\subsection{RL with Jev-Reweighted Replay}
\label{sec:jev-sampling}




Beyond guiding exploration, Jev can also judge which collected experience to learn from. Uniform replay gives all stored transitions the same sampling probability \citep{lin1992self}, although their usefulness for learning may differ.   Prioritized experience replay assigns higher probabilities to transitions with larger TD errors \citep{schaul2016per}. Here, we take the Jev model as an external importance judge, using the task description and replay context to inform which transitions deserve more updates. We instantiate this use as \emph{Jev-Reweighted Replay}, which uses the \ptag{Jev as Score Rater} position of Sec.~\ref{sec:roles-list}, and keep both the action rule and the task reward
unchanged so that only the replay distribution moves.

For a stored transition $z_i=(s_i,a_i,r_i,s'_i,d_i)$, Jev receives its structured before/after
description $\tilde z_i$ together with the current replay evidence $B_t$. Its rating becomes the
sampling probability directly,
\begin{equation}
 I_i=R_J((\tilde z_i,B_t);u_{\mathrm{imp}}),\qquad
 P_J(i)=\frac{\max(I_i,10^{-3})}{\sum_{j\in\mathcal D}\max(I_j,10^{-3})},
 \label{eq:jev-priority}
\end{equation}
where the floor retains support for every stored transition.

\vspace{-0.3cm}
\paragraph{\yhead{Jev-Reweighted Replay.}}
We instantiate two variants according to how importance is judged. \emph{Intuitive importance} asks for a transition's
learning importance without prescribing what makes it important, and \emph{success importance}
asks for its contribution to the final task objective, including prerequisites and informative
failures. Both receive the same inputs, so only the instruction differs
(Table~\ref{tab:other-prompts}). New transitions are rated on insertion, and up to eight stored transitions are re-rated every 100 environment steps using the current replay context. The critic minimizes its squared TD loss on batches drawn from $P_J$. In addition, no importance-sampling correction is applied, so the ratings deliberately change how transitions
contribute to the expected training loss. A judgment can carry useful task information and still
allocate updates poorly for the learner's current needs.

A fourth channel, subgoal selection, asks for a judgment whose completion requires a
sequence of actions rather than a single transition. We examine it separately in Appendix~\ref{app:subgoal}.

\section{Experiments}
\label{sec:experiments}

\providecommand{\JevFinding}[2]{%
  \par\smallskip
  \begingroup
  \setlength{\fboxsep}{6pt}%
  \setlength{\fboxrule}{0.45pt}%
  \noindent\fcolorbox[rgb]{0.46,0.31,0.61}{0.97,0.96,0.985}{%
    \parbox{\dimexpr\linewidth-2\fboxsep-2\fboxrule\relax}{%
      \textbf{Finding #1.}\ #2%
    }%
  }%
  \endgroup
  \par\smallskip
}

We evaluate Jev in two steps: first as a decision maker, then as a component of learning. RQ1--RQ2 examine direct control and its dependence on the information interface; RQ3--RQ5 follow the three channels of Sec.~4: policy reference, exploration, and replay. Subgoal guidance is examined separately in Appendix~\ref{app:subgoal}.

\paragraph{Experimental setup.}
We use nine MiniGrid tasks: DoorKey at three sizes, Empty, FourRooms, two MultiRoom variants, SimpleCrossing, and LavaGap, and three Atari games: Seaquest, Freeway, and SpaceInvaders. MiniGrid students use MLPs over native local observations; Atari students use CNNs over four stacked grayscale frames. Implementations build on CleanRL~\citep{huang2021cleanrl}, with DQN as the common task baseline and TD-error prioritized experience replay (PER)~\citep{schaul2016per} as an additional replay baseline. Jev remains frozen and receives structured observation descriptions, task rules, and query-specific history. These inputs may carry information unavailable to the student. FourRooms and MultiRoom-N4 action queries additionally include observed-map memory and engineered navigation features. Configurations and input examples appear in Appendices~\ref{app:experiment-details} and~\ref{app:complete-inputs}. Main results measure success at 10k MiniGrid steps and native return at 50k Atari decisions, with four emulator frames per decision. Learned policies are evaluated without Jev or auxiliary rewards on 50 fixed MiniGrid seeds or five fixed Atari seeds. Learning results report means and SEM across three training seeds; fixed-experience diagnostics use 100 evaluation seeds. Appendix~\ref{app:experiment-details} provides learning curves and detailed protocols. 


\subsection{RQ1: How capable is Jev as a zero-shot policy?}
\label{sec:exp_rq1}

\begin{figure}[!htbp]
  \centering
  \includegraphics[width=\linewidth]{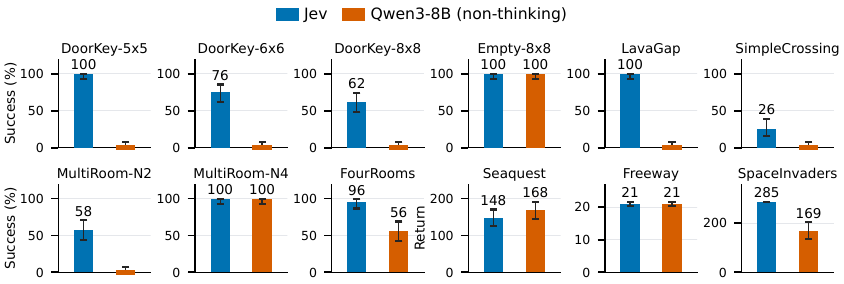}
  \vspace{-0.5cm}
  \caption{\textbf{Direct control with matched structured inputs.}
  Jev and non-thinking Qwen3-8B use the same evaluation seeds and task-specific input construction. MiniGrid: success on 50 episodes with 95\% Wilson intervals. Atari: mean native return over five complete games with SEM. A colored horizontal stroke denotes zero successes. FourRooms and MultiRoom-N4 use the full-navigation input; the other tasks use their direct-Jev input interface.}
  \label{fig:matched-models}
\end{figure}

We compare Jev with Qwen3-8B under the same structured-input pipeline, task questions, legal-action descriptions, and evaluation seeds. Qwen uses non-thinking, temperature-zero decoding constrained to a JSON action label; Jev executes its most probable action. FourRooms and MultiRoom-N4 use the shared full-navigation interface. Jev matches or exceeds the tested Qwen configuration on all nine MiniGrid tasks, whereas Atari comparisons are mixed (Figure~\ref{fig:matched-models}). The LavaGap action audit further shows useful discrimination between goal and hazard cues, although direct-policy success varies across tasks. Thus, reliable local choices and reliable completion of a long-horizon task remain different achievements.

\vspace{-0.5cm}
\JevFinding{1}{%
Jev supplies a useful zero-shot action prior, but not a uniformly reliable controller. Its local decision capability motivates its use in learning without establishing reliable completion of a long-horizon task.
}

\subsection{RQ2: How much does the information interface matter?}
\label{sec:exp_rq2}

We keep Jev fixed and remove candidate routes, path distances, and progress/priority hints on four tasks, retaining identical observational memory and task rules. This isolates navigation features rather than memory itself. DoorKey-8$\times$8 and MultiRoom-N2 use shared-memory prompts distinct from the local-input comparison in RQ1. Removing these features sharply reduces success on FourRooms and MultiRoom-N4, while DoorKey-8$\times$8 and MultiRoom-N2 remain largely solvable (Table~\ref{tab:navigation}). Routes also lengthen on jointly solved instances.

\begin{table}[t]
  \centering
  \small
  \setlength{\tabcolsep}{7pt}
  \caption{\textbf{Task-dependent contribution of engineered navigation.}
  Pure-Jev performance over 50 paired environment seeds. Episode length is
  averaged over instances solved in both conditions.}
  \vspace{-0.3cm}
  \label{tab:navigation}
  \scalebox{0.8}{
  \begin{tabular}{lcc|cc}
    \toprule
    & \multicolumn{2}{c|}{Success (\%)} &
      \multicolumn{2}{c}{Episode length} \\
    \textbf{Environment}
      & \textbf{With navigation} & \textbf{Without navigation}
      & \textbf{With navigation} & \textbf{Without navigation} \\
    \midrule
    DoorKey-8$\times$8
      & \textbf{100} & \textbf{100} & \textbf{17.2} & 25.6 \\
    MultiRoom-N2
      & \textbf{100} & 90 & \textbf{6.6} & 11.2 \\
    FourRooms
      & \textbf{96} & 40 & \textbf{21.7} & 31.3 \\
    MultiRoom-N4
      & \textbf{100} & 38 & \textbf{32.9} & 59.2 \\
    \bottomrule
  \end{tabular}
  }
\end{table}
\vspace{-0.2cm}
\JevFinding{2}{%
The relevant capability is \emph{the model's decision under a specified information interface}. High success with engineered navigation features demonstrates use of supplied route structure, not its independent discovery by Jev.
}

\subsection{RQ3: Can Jev guide policy learning?}
\label{sec:exp_rq3}

\begin{figure}[!htbp]
  \centering
  \includegraphics[width=\linewidth]{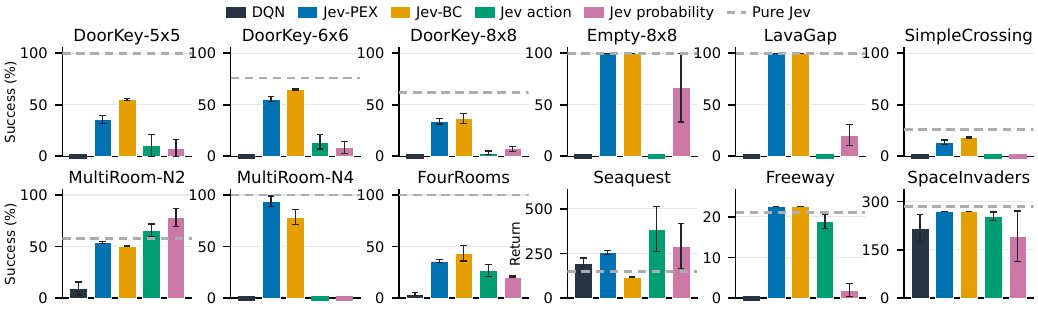}
  \caption{\textbf{RL with Jev Reference.}
  Endpoint performance on all twelve environments: MiniGrid at 10k steps (success percentage) and Atari at 50k decisions (return). Error bars are SEM across training seeds. A colored horizontal stroke denotes a measured zero. Hatched bars show independent pure-Jev evaluations without training-seed error bars.}
  \label{fig:reference-bars}
\end{figure}

We compare Jev-BC, Jev-PEX, and Jev-Mixed Sampling, which use the action prior for imitation, critic-mediated guidance, and experience collection, respectively. Pure-Jev evaluations use the corresponding reference-method inputs. These uses improve learning on multiple tasks, including settings where DQN remains unsuccessful within the budget (Figure~\ref{fig:reference-bars}). Their ranking nevertheless varies: Seaquest benefits more from reference-guided collection than from imitation. 
We also find teacher--student agreement on fixed logged states increases substantially during training (Figure~\ref{fig:teacher-uptake} in Appendix~\ref{sec:analysis-experiments}), especially when the goal is visible. The student acquires local action knowledge without necessarily reproducing the teacher's behavior over an entire task.

\vspace{-0.3cm}
\JevFinding{3}{%
Jev's frozen action prior can improve policy learning and sometimes enable students to outperform direct Jev control. The teacher thus provides guidance rather than a performance ceiling, with the gains depending on how its prior is integrated into learning.
}

\subsection{RQ4: Can Jev improve exploration?}
\label{sec:exp_rq4}

\begin{figure}[!htbp]
  \centering
  \includegraphics[width=0.8\linewidth]{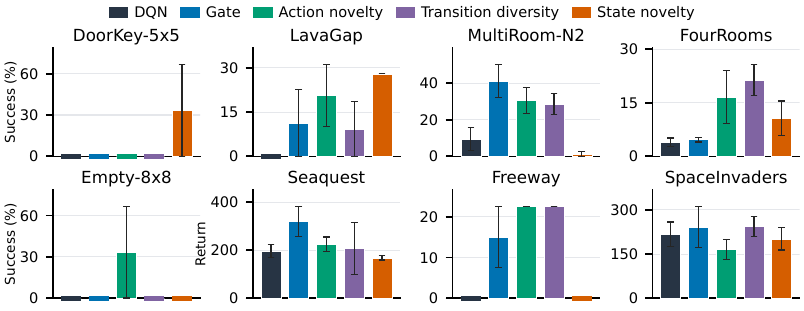}
  \vspace{-0.3cm}
  \caption{\textbf{RL with Jev Exploration.}
  MiniGrid 10k and Atari 50k endpoints, mean$\pm$SEM across three training seeds.  Tasks with zero performance for all four methods are omitted.}
  \label{fig:exploration-bars}
\end{figure}

Jev-Gate and the three Jev-Bonus variants improve selected tasks without asking Jev to choose a task-preferred action, but no tested configuration is uniformly best (Figure~\ref{fig:exploration-bars}). We separate the contributions of collected experience and auxiliary reward through two fixed-experience controls.

First, we retrain identical learners on each method's recorded experience using only environment rewards. State-novelty data improves LavaGap learning despite having the same mean spatial coverage as DQN data (Figure~\ref{fig:data-quality} in Appendix~\ref{sec:analysis-experiments}). Part of the exploration gain therefore survives removal of the auxiliary reward and lies in the learning opportunities present in the collected experience. Second, we fix the experience and replace each Jev state score with the mean for its visitation bin and 2k-step collection window. Original scores outperform these group means on two of three tasks (Figure~\ref{fig:state-scores} in Appendix~\ref{sec:analysis-experiments}). This identifies useful reward distinctions beyond the coarse count summary, rather than a general advantage over count-based exploration.

\JevFinding{4}{%
Jev can collect more useful experience, not just expand coverage. Its state-dependent bonuses add reward information beyond coarse visitation summaries.
}

\subsection{RQ5: Can Jev improve replay?}
\label{sec:exp_rq5}
\begin{figure}[!htbp]
  \centering
  \includegraphics[width=0.8\linewidth]{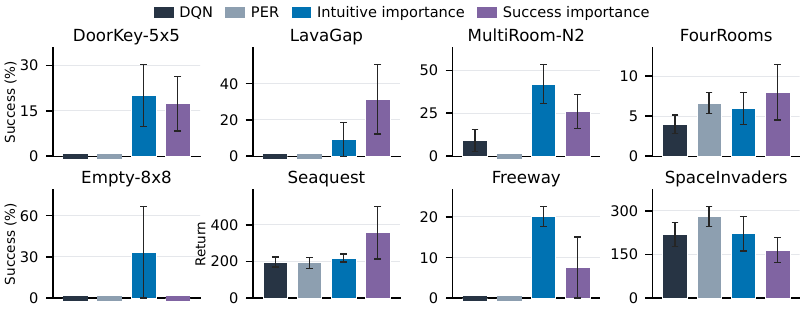}
  \vspace{-0.3cm}
  \caption{\textbf{RL with Jev-Reweighted Replay.}
  MiniGrid 10k and Atari 50k endpoints, mean$\pm$SEM across three training seeds. Tasks with zero performance for all four methods are omitted.}
  \label{fig:sampling-bars}
\end{figure}

We compare replay guided by Jev's intuitive and success-oriented ratings with uniform replay and PER, keeping DQN's action rule and task reward unchanged. Jev improves selected tasks but does not consistently outperform PER, and several tasks remain unsolved within the budget (Figure~\ref{fig:sampling-bars}). To isolate replay effects from changes in data collection, we fix each sampler's experience and rating history. We compare the original weights with uniform replay and a reward-class control that averages weights separately for positive- and non-positive-reward transitions. This preserves each class's total sampling probability but treats transitions equally within each class, testing whether Jev helps beyond identifying rewarded transitions. On MultiRoom-N2, intuitive weights outperform both controls; success-oriented weights, tested on separately collected data, outperform the reward-class control but underperform uniform replay (Figure~\ref{fig:sampling-analysis} in Appendix~\ref{sec:analysis-experiments}). Thus, Jev can improve replay beyond simply favoring rewarded transitions, but following its ratings is not always better than uniform sampling.


\JevFinding{5}{%
Jev helps select useful experience for replay, but success-oriented ratings do not consistently outperform intuitive ratings.

}


\section{Conclusion}
\label{sec:conclusion}

We examined which RL roles Jev can fulfill and developed algorithms that use the frozen model as a policy reference, exploration judge, and replay rater. Experiments on nine MiniGrid tasks and three Atari games show that these uses can improve early learning, sometimes enabling students to outperform direct Jev control while acting independently at evaluation. These benefits depend on the task, information interface, and how Jev's judgments enter learning. Our findings demonstrate that a decision model need not be a reliable standalone controller to improve RL training, motivating guidance that adapts to the learner's needs.

\subsubsection*{Reproducibility Statement}
Appendix~\ref{app:experiment-details} specifies evaluation seeds, budgets,
learner settings, and fixed-experience controls.
Appendix~\ref{app:complete-inputs} documents query instructions, answer choices,
state representations, and memory. The accompanying source files contain the
figure inputs and plotting scripts. Direct-model trajectories are retained for
observation and action replay checks. Jev results are tied to the recorded API
version and input construction.

\subsubsection*{Ethics Statement}
Our experiments are confined to MiniGrid and Atari environments. Jev's judgments are task- and input-dependent, and improved benchmark performance does not establish the safety of learned policies in real-world settings. Any safety-critical application should undergo independent evaluation, with appropriate safeguards and human oversight. 

\subsubsection*{AI Use Statement}
AI coding and language tools assisted with experiment implementation and manuscript polishing.
Jev and Qwen3-8B are also experimental subjects; their roles and evaluation
interfaces are specified separately in Secs.~\ref{sec:method}
and~\ref{sec:experiments}.

\bibliography{references}

@inproceedings{chen2021decision,title={{Decision Transformer}: Reinforcement Learning via Sequence Modeling},author={Chen, Lili and Lu, Kevin and Rajeswaran, Aravind and Lee, Kimin and Grover, Aditya and Laskin, Michael and Abbeel, Pieter and Srinivas, Aravind and Mordatch, Igor},booktitle={Advances in Neural Information Processing Systems},volume={34},pages={15084--15097},year={2021}}

@inproceedings{yao2023react,title={{ReAct}: Synergizing Reasoning and Acting in Language Models},author={Yao, Shunyu and Zhao, Jeffrey and Yu, Dian and Du, Nan and Shafran, Izhak and Narasimhan, Karthik and Cao, Yuan},booktitle={International Conference on Learning Representations},year={2023}}

@article{feng2024nlrl,title={Natural Language Reinforcement Learning},author={Feng, Xidong and Liu, Bo and Song, Yan and Fu, Haotian and Wan, Ziyu and Koushik, Girish A. and Hu, Zhiyuan and Yang, Mengyue and Wen, Ying and Wang, Jun},journal={arXiv preprint arXiv:2411.14251},year={2024}}

@inproceedings{ma2024eureka,title={Eureka: Human-Level Reward Design via Coding Large Language Models},author={Ma, Yecheng Jason and Liang, William and Wang, Guanzhi and Huang, De-An and Bastani, Osbert and Jayaraman, Dinesh and Zhu, Yuke and Fan, Linxi and Anandkumar, Anima},booktitle={International Conference on Learning Representations},year={2024}}

@inproceedings{rocamonde2024vlmrm,title={Vision-Language Models are Zero-Shot Reward Models for Reinforcement Learning},author={Rocamonde, Juan and Montesinos, Victoriano and Nava, Elvis and Perez, Ethan and Lindner, David},booktitle={International Conference on Learning Representations},year={2024}}

@inproceedings{yao2023tree,title={Tree of Thoughts: Deliberate Problem Solving with Large Language Models},author={Yao, Shunyu and Yu, Dian and Zhao, Jeffrey and Shafran, Izhak and Griffiths, Thomas L. and Cao, Yuan and Narasimhan, Karthik},booktitle={Advances in Neural Information Processing Systems},volume={36},year={2023}}

@misc{typesafe2026jev,title={Introducing {System One} Models \& {Jev}},author={{TypeSafe AI}},howpublished={TypeSafe AI blog, \url{https://typesafe.ai/blog/introducing-system-one-models-and-jev}},year={2026},note={Vendor material; no peer-reviewed description of {RLCD} exists}}

@article{rafe2026jev,title={Calibrated Decisions at Scale: Converting Police Crash Narratives into Probabilistic Crash Variables with a {System One Model} ({Jev})},author={Rafe, Amir and Das, Subasish},journal={arXiv preprint arXiv:2609.24052},year={2026}}

@article{huang2026radiology,title={Can {Jev} Judge Radiology Reports? Evaluating a {System One Model} for Clinical Factuality},author={Huang, Jiaju and Yang, Hao and Ma, Xinyu and Liang, Xinglong and Cai, Kunyan and Ma, Junqiang and Chen, Shaobin and Sun, Yue and Tan, Tao},journal={arXiv preprint arXiv:2609.27607},year={2026}}

@article{robitza2026video,title={{JEVQA} - Video Quality from Metadata, Bitstream, and Pixel Features with a General-Purpose Decision Model},author={Robitza, Werner},journal={arXiv preprint arXiv:2609.24395},year={2026}}

@book{sutton2018reinforcement,title={Reinforcement Learning: An Introduction},author={Sutton, Richard S. and Barto, Andrew G.},edition={2},publisher={MIT Press},year={2018}}

@misc{typesafe2026docs,title={{TypeSafe AI} Documentation},author={{TypeSafe AI}},howpublished={\url{https://docs.typesafe.ai/}},year={2026}}

@article{yang2023foundation,title={Foundation Models for Decision Making: Problems, Methods, and Opportunities},author={Yang, Sherry and Nachum, Ofir and Du, Yilun and Wei, Jason and Abbeel, Pieter and Schuurmans, Dale},journal={arXiv preprint arXiv:2303.04129},year={2023}}

@article{cao2024survey,title={Survey on Large Language Model-Enhanced Reinforcement Learning: Concept, Taxonomy, and Methods},author={Cao, Yuji and Zhao, Huan and Cheng, Yuheng and Shu, Ting and Chen, Yue and Liu, Guolong and Liang, Gaoqi and Zhao, Junhua and Yan, Jinyue and Li, Yun},journal={IEEE Transactions on Neural Networks and Learning Systems},volume={36},number={6},pages={9737--9757},year={2025},note={arXiv:2404.00282}}

@inproceedings{wang2024rlvlmf,title={{RL-VLM-F}: Reinforcement Learning from Vision Language Foundation Model Feedback},author={Wang, Yufei and Sun, Zhanyi and Zhang, Jesse and Xian, Zhou and Biyik, Erdem and Held, David and Erickson, Zackory},booktitle={Proceedings of the 41st International Conference on Machine Learning},series={Proceedings of Machine Learning Research},volume={235},pages={51484--51501},year={2024},publisher={PMLR}}

@inproceedings{bougie2025exploring,title={Exploring Beyond Curiosity Rewards: Language-Driven Exploration in {RL}},author={Bougie, Nicolas and Watanabe, Narimasa},booktitle={Proceedings of the 16th Asian Conference on Machine Learning},series={Proceedings of Machine Learning Research},volume={260},pages={127--142},year={2025},publisher={PMLR}}

@inproceedings{klissarov2025maestromotif,title={{MaestroMotif}: Skill Design from Artificial Intelligence Feedback},author={Klissarov, Martin and Henaff, Mikael and Raileanu, Roberta and Sodhani, Shagun and Vincent, Pascal and Zhang, Amy and Bacon, Pierre-Luc and Precup, Doina and Machado, Marlos C. and D'Oro, Pierluca},booktitle={International Conference on Learning Representations},year={2025},note={arXiv:2412.08542}}

@inproceedings{zhou2024llmteacher,title={Large Language Model as a Policy Teacher for Training Reinforcement Learning Agents},author={Zhou, Zihao and Hu, Bin and Zhao, Chenyang and Zhang, Pu and Liu, Bin},booktitle={Proceedings of the Thirty-Third International Joint Conference on Artificial Intelligence (IJCAI)},pages={5671--5679},year={2024},note={arXiv:2311.13373}}

@inproceedings{yan2025efficient,title={Efficient Reinforcement Learning with Large Language Model Priors},author={Yan, Xue and Song, Yan and Feng, Xidong and Yang, Mengyue and Zhang, Haifeng and Bou Ammar, Haitham and Wang, Jun},booktitle={International Conference on Learning Representations},year={2025},note={arXiv:2410.07927}}

@article{zhang2024linvit,title={How Can {LLM} Guide {RL}? A Value-Based Approach},author={Zhang, Shenao and Zheng, Sirui and Ke, Shuqi and Liu, Zhihan and Jin, Wanxin and Yuan, Jianbo and Yang, Yingxiang and Yang, Hongxia and Wang, Zhaoran},journal={arXiv preprint arXiv:2402.16181},year={2024}}

@article{ma2026jevstar,title={{JEV-Star}: Fast, Low-Cost {StarCraft II} Control with Language-Model Planning},author={Ma, Weiyu and Zhao, Liangbing and Zeng, Yongcheng and Zhao, Jian},journal={arXiv preprint arXiv:2609.27331},year={2026}}

@article{wu2026reflex,title={{REFLEX} with {Jev} for Efficient Selective Control in {LLM} Agents},author={Wu, Tiantong and Lim, Wei Yang Bryan},journal={arXiv preprint arXiv:2609.26532},year={2026}}

@article{ibrahim2026evaluating,title={Evaluating Decision Models for Text Annotation in Computational Social Science},author={Ibrahim, Hazem and Zaki, Yasir},journal={arXiv preprint arXiv:2609.24574},year={2026}}

@article{zhang2026samescores,title={Same Scores, Different Decisions: Evaluating {JEV} and Language Models for Legal Document Understanding},author={Zhang, Fan and Chen, Yankai and Xie, Zhuohan and Zhou, Yixi and Peng, Sijia and Fan, Lei and Ji, Xinhua and Zheng, Cunyuan and Shan, Huangyong and Yu, Philip S. and Liu, Xue and Chen, Yu and Nakov, Preslav and He, Songwei},journal={arXiv preprint arXiv:2609.27678},year={2026}}

@article{deng2026jevscidecisions,title={{Jev} for Scientific Decisions: Evaluating Semantic Choices and Their Consequences},author={Deng, Boyuan and Fan, Shuyi and Zhang, Hongyang and Xie, Xinhong},journal={arXiv preprint arXiv:2609.24965},year={2026}}

@article{jiang2026jevmem,title={{Jev-Mem}: {System-One-Controlled} Agentic Memory for Efficient {AI} Agents},author={Jiang, Dongming and Li, Yi and Li, Bingzhe},journal={arXiv preprint arXiv:2609.23986},year={2026}}

@article{li2026fastintent,title={Fast Intent-Driven Service Orchestration with {Jev} for {6G} Edge Networks},author={Li, Delong and Wang, Xu and Gong, Haochen and Lang, Rui and Yu, Guangsheng},journal={arXiv preprint arXiv:2609.23136},year={2026}}

@article{li2026replacing,title={Replacing Large Language Models with {Jev} Decision Models for Low-Latency Edge Service Orchestration},author={Li, Delong and Wang, Xu and Gong, Haochen and Lang, Rui and Yu, Guangsheng},journal={arXiv preprint arXiv:2609.22753},year={2026}}

@article{li2026kite,title={{KITE}: Scaling {Jev} Population Experiments with Sparse Flagship Calibration},author={Li, Hengyu},journal={arXiv preprint arXiv:2609.27535},year={2026}}

@article{ren2026openjev,title={{Open-Jev} Judgments on {CallScreenBench}: Calibrated One-Pass Scam Screening with a Small Language Model},author={Ren, Simiao and Zewde, Kidus and Shen, Xingyu and Zhou, Yuchen and Ng, Dennis and Raj, Ankit and Duong, Tommy and Zhang, Yuxin and Tiangratanakul, Neo},journal={arXiv preprint arXiv:2609.23959},year={2026}}

@article{li2026jevjudge,title={{JEV}-as-a-Judge: Accept When Confident, Escalate When Unsure},author={Li, Yubo and Miao, Yidi and Krishnan, Ramayya and Padman, Rema},journal={arXiv preprint arXiv:2609.26550},year={2026}}

@article{sun2026typesafe,title={Type-Safe Is Not Error-Free: A Constrained Decision Head Follows the Option Name, Not the Rubric Bound to It},author={Sun, Yu and Xu, Junhao and Shi, Jiajia and Yang, Zijin},journal={arXiv preprint arXiv:2609.26758},year={2026}}

@article{wu2026decisionhijacking,title={Decision Hijacking: Prompt Injection Attacks on {Jev}'s Typed Probabilistic Decisions},author={Wu, Tiantong and Lim, Wei Yang Bryan},journal={arXiv preprint arXiv:2609.28613},year={2026}}

@article{yu2026visualjev,title={Visual {Jev}: Accurate and Efficient Decisions from Shared Visual Context},author={Yu, Guanxu and Yao, Yuhang},journal={arXiv preprint arXiv:2609.25845},year={2026}}

@article{ye2026numericjev,title={{NumericJev}: {Jev}-like {LLM} Numerical Decoding with Multiway Decision Trees},author={Ye, Weiwei and Liu, Hangchen and Jiang, Renhe},journal={arXiv preprint arXiv:2609.28587},year={2026}}

@article{cheng2026thisthat,title={this-that-model-1.0: A typed decision model that decides in 30 ms, for a millionth of a cent},author={Cheng, Zehua and Dai, Wei and Sun, Jiahao},journal={arXiv preprint arXiv:2609.23886},year={2026}}

@inproceedings{zhang2023pex,
 title={Policy Expansion for Bridging Offline-to-Online Reinforcement Learning},
 author={Zhang, Haichao and Xu, Wei and Yu, Haonan},
 booktitle={International Conference on Learning Representations},year={2023},
 url={https://arxiv.org/abs/2302.00935}}

@article{huang2021cleanrl,
 title={{CleanRL}: High-quality Single-file Implementations of Deep Reinforcement Learning Algorithms},
 author={Huang, Shengyi and Dossa, Rousslan Fernand Julien and Ye, Chang and Braga, Jeff and Chakraborty, Dipam and Mehta, Kinal and Ara{\'u}jo, Jo{\~a}o G. M.},
 journal={Journal of Machine Learning Research},volume={23},number={274},pages={1--18},year={2022},
 url={https://jmlr.org/papers/v23/21-1342.html}}

@inproceedings{schaul2016per,
 title={Prioritized Experience Replay},author={Schaul, Tom and Quan, John and Antonoglou, Ioannis and Silver, David},
 booktitle={International Conference on Learning Representations},year={2016},url={https://arxiv.org/abs/1511.05952}}

@article{schulman2017ppo,
 title={Proximal Policy Optimization Algorithms},
 author={Schulman, John and Wolski, Filip and Dhariwal, Prafulla and Radford, Alec and Klimov, Oleg},
 journal={arXiv preprint arXiv:1707.06347}, year={2017}
}

@inproceedings{burda2019rnd,
 title={Exploration by Random Network Distillation},
 author={Burda, Yuri and Edwards, Harrison and Storkey, Amos and Klimov, Oleg},
 booktitle={International Conference on Learning Representations}, year={2019}
}

@article{pomerleau1991,
 title={Efficient Training of Artificial Neural Networks for Autonomous Navigation},
 author={Pomerleau, Dean A.},
 journal={Neural Computation}, volume={3}, number={1}, pages={88--97}, year={1991},
 doi={10.1162/neco.1991.3.1.88}
}

@inproceedings{ross2011dagger,
 title={A Reduction of Imitation Learning and Structured Prediction to No-Regret Online Learning},
 author={Ross, St{\'e}phane and Gordon, Geoffrey J. and Bagnell, J. Andrew},
 booktitle={Proceedings of the 14th International Conference on Artificial Intelligence and Statistics (AISTATS)},
 series={Proceedings of Machine Learning Research}, volume={15}, pages={627--635}, year={2011}
}

@article{hussein2017imitation,
 title={Imitation Learning: A Survey of Learning Methods},
 author={Hussein, Ahmed and Gaber, Mohamed Medhat and Elyan, Eyad and Jayne, Chrisina},
 journal={ACM Computing Surveys}, volume={50}, number={2}, pages={21:1--21:35}, year={2017}
}

@inproceedings{pathak2017curiosity,
 title={Curiosity-driven Exploration by Self-supervised Prediction},
 author={Pathak, Deepak and Agrawal, Pulkit and Efros, Alexei A. and Darrell, Trevor},
 booktitle={Proceedings of the 34th International Conference on Machine Learning},
 series={Proceedings of Machine Learning Research}, volume={70}, pages={2778--2787}, year={2017}
}

@inproceedings{bellemare2016unifying,
 title={Unifying Count-Based Exploration and Intrinsic Motivation},
 author={Bellemare, Marc G. and Srinivasan, Sriram and Ostrovski, Georg and Schaul, Tom and Saxton, David and Munos, R{\'e}mi},
 booktitle={Advances in Neural Information Processing Systems},
 pages={1471--1479}, year={2016}
}

@article{lin1992self,
 title={Self-improving Reactive Agents Based on Reinforcement Learning, Planning and Teaching},
 author={Lin, Long-Ji},
 journal={Machine Learning}, volume={8}, number={3--4}, pages={293--321}, year={1992}
}

@inproceedings{chevalier2023minigrid,
 title={{Minigrid} \& {Miniworld}: Modular \& Customizable Reinforcement Learning Environments for Goal-Oriented Tasks},
 author={Chevalier-Boisvert, Maxime and Dai, Bolun and Towers, Mark and Perez-Vicente, Rodrigo and Willems, Lucas and Lahlou, Salem and Pal, Suman and Castro, Pablo Samuel and Terry, Jordan},
 booktitle={Advances in Neural Information Processing Systems}, volume={36},
 note={Datasets and Benchmarks Track}, year={2023},
 url={https://arxiv.org/abs/2306.13831}
}

@article{bellemare2013ale,
 title={The {Arcade Learning Environment}: An Evaluation Platform for General Agents},
 author={Bellemare, Marc G. and Naddaf, Yavar and Veness, Joel and Bowling, Michael},
 journal={Journal of Artificial Intelligence Research}, volume={47}, pages={253--279}, year={2013},
 doi={10.1613/jair.3912}
}

@article{yang2025qwen3,
 title={{Qwen3} Technical Report},
 author={{Qwen Team}},
 journal={arXiv preprint arXiv:2505.09388}, year={2025}
}
\bibliographystyle{iclr2027_conference}

\clearpage
\begin{center}
    {\Huge\bfseries\textcolor{appendixpurple}{Appendix Contents}}
\end{center}
\vspace{1.0em}

\begin{tcolorbox}[
    colback=appendixlightpurple,
    colframe=appendixpurple,
    boxrule=0.8pt,
    arc=2pt,
    left=1.1em,
    right=1.1em,
    top=1.0em,
    bottom=1.0em
]
\begingroup
\renewcommand{\arraystretch}{1.55}
\setlength{\tabcolsep}{0pt}
\begin{tabularx}{\textwidth}{@{}p{3.2em}@{\hspace{1.4em}}Xr@{}}
\appentry{\ref*{app:jev}}{Jev: Experimental Interface}{app:jev}
\appentry{\ref*{app:jev-lit}}{Jev: An Early Literature Snapshot}{app:jev-lit}
\appentry{\ref*{app:experiment-details}}{Experimental Details and Learning Curves}{app:experiment-details}
\appentry{\ref*{sec:analysis-experiments}}{Analysis Experiments}{sec:analysis-experiments}
\appentry{\ref*{app:subgoal}}{Jev for Subgoal Guidance}{app:subgoal}
\appsubentry{\ref*{app:subgoal-channel}}{Channel}{app:subgoal-channel}
\appsubentry{\ref*{app:subgoal-results}}{Results}{app:subgoal-results}
\appentry{\ref*{app:complete-inputs}}{Structured Input Examples}{app:complete-inputs}
\end{tabularx}
\endgroup
\end{tcolorbox}

\clearpage

\newpage
\appendix
\onecolumn

\section{Jev: Experimental Interface}
\label{app:jev}

The experiments use \texttt{jev-1.13.0} as a fixed external decision model.
Table~\ref{tab:jev-params} records the interface settings used in this study.
These are experimental configurations, rather than measurements of the API's
maximum capacity or vendor-reported performance.

\begin{table}[h]
 \centering
 \caption{Interface choices used in the experiments.}
 \label{tab:jev-params}
 \begin{tabular}{ll}
  \toprule
  Component & Experimental setting\\
  \midrule
  API model identifier & \texttt{jev-1.13.0}\\
  Query primitive & \texttt{choice}\\
  MiniGrid action options & Seven native primitive actions\\
  Atari action options & Seaquest 18; Freeway 3; SpaceInvaders 6\\
  Exploration-gate options & \texttt{GREEDY}, \texttt{RANDOM}\\
  Rating options & Five ordered levels and \texttt{unknown}\\
  Rating values & $0,.25,.5,.75,1$; unknown contributes zero\\
  Subgoal options & Observed, task-specific candidate milestones\\
  \bottomrule
 \end{tabular}
\end{table}

Action preference and numerical rating use the same output primitive with
different questions. In a rating query, the expectation in the rating of
Sec.~\ref{sec:jev-exploration} can fall between rubric levels. Its numerical spacing
comes from our assigned auxiliary scale, not from calibration to task returns.
A question about action quality, a question about novelty, and a question about
replay importance therefore require distinct instructions even when all return
probabilities over named options. Full query distinctions and state examples are
in Appendix~\ref{app:complete-inputs}.

\section{Jev: An Early Literature Snapshot}
\label{app:jev-lit}

This appendix groups the additional Jev-related work cited in our bibliography
by application. It provides context for the decision-model interface rather than
a quantitative comparison with the control experiments in Sec.~\ref{sec:experiments}.

\textbf{Documents and text.} Coding of police crash narratives \citep{rafe2026jev},
clinical factuality in radiology reports \citep{huang2026radiology}, semantic relation
selection in scientific workflows \citep{deng2026jevscidecisions}, and legal contract
understanding \citep{zhang2026samescores}.

\textbf{Agent and system infrastructure.} A control plane for agent memory
\citep{jiang2026jevmem}, selective control of a stronger model \citep{wu2026reflex},
and intent interpretation for service orchestration in edge networks
\citep{li2026fastintent,li2026replacing}.

\textbf{Simulation and annotation infrastructure.} Million-agent population simulation
that queries the model once per unique state \citep{li2026kite}, and one-pass scam-call
screening \citep{ren2026openjev}.

\textbf{Evaluation.} A multi-task comparison against nineteen language models
\citep{ibrahim2026evaluating}, a judge cascade with confidence-based escalation
\citep{li2026jevjudge}, a study showing that rebinding option names to fixed rubrics
flips or reverses decisions \citep{sun2026typesafe}, and prompt-injection attacks on
typed probabilistic decisions \citep{wu2026decisionhijacking}.

Several further papers use the typed-decision interface without evaluating Jev itself:
shared visual context encoding \citep{yu2026visualjev}, training-free numeric decoding
\citep{ye2026numericjev}, and a competing open decision model that reports Jev as a
baseline \citep{cheng2026thisthat}.

These applications motivate studying decisions as reusable components in agent
systems. Our experiments examine a complementary question: how a fixed model's
judgments affect a separately trained RL learner through explicitly defined
channels.

\section{Experimental Details and Learning Curves}
\label{app:experiment-details}

\paragraph{Evaluation and uncertainty.}
MiniGrid evaluations use seeds 700000--700049 and a greedy student. Atari uses
seeds 700000--700004 and the student's $\epsilon=.01$ evaluation policy. Rewards
in every evaluation are the native task rewards, without the auxiliary training
bonus. Jev-PEX/Jev-BC use five training seeds for Seaquest, DoorKey-5/8, MultiRoom-N2,
SimpleCrossing, and LavaGap, and three elsewhere; the other plotted learning
methods use three. Error bars and shaded bands are SEM over training seeds, not
uncertainty over individual evaluation episodes. Fixed-experience analyses use
100 evaluation seeds, 900000--900099, and three data/training seeds.

\paragraph{Input construction.}
MiniGrid teachers receive descriptions derived from native local observations.
DoorKey inputs identify key/door state and interaction rules; LavaGap describes
lava and safe movement; room-navigation inputs describe passages and doors.
FourRooms and MultiRoom-N4 action teachers additionally retain observed-map memory
and engineered route advice. The state-novelty query omits those action proposals
and uses the local observation, visit counts, and bounded observation history. Table~\ref{tab:query-memory} distinguishes
recent-state and recent-transition records from cumulative counts, including
their lengths and reset rules.
Atari descriptions use RGB-detected objects and estimated motion: oxygen, divers and
enemies for Seaquest; vehicles and the chicken for Freeway; and the cannon, aliens,
projectiles and shields for SpaceInvaders. MiniGrid has seven native actions;
Seaquest, Freeway and SpaceInvaders have 18, 3 and 6, respectively. The recorded
MiniGrid-MultiRoom-N4-S5 registration generates six rooms with a 120-step limit;
we retain the experiment identifier as its short label.

These input adapters are also the common basis of the model comparison.
The vLLM service identifies its deployed model as Qwen3-8B. We serialize the same
structured situation, question, and legal-action descriptions used by Jev, and
request a JSON action label with temperature zero, a 48-token output limit, and
thinking disabled. The comparison covers nine MiniGrid environments (seeds
700000--700049) and three Atari games (seeds 700000--700004), with the same
native episode limits for both models. MiniGrid has seven primitive choices;
Atari retains each game's original action set. Stored Qwen trajectories are replayed to verify every
observation, reconstructed input, action mapping, reward, and termination.

To vary the input rather than the model, the navigation ablation uses a shared raw-memory prompt in both arms; its full
control is distinct from the original direct-policy prompt in the main reference
figure. The four-task ablation uses 50 paired seeds per task. DoorKey-8 adds
observed-map routes to keys, usable doors, goals, and frontiers; the no-navigation
arm retains identical task rules and raw memory. MultiRoom-N2 uses two-room
rules with the same route construction as MultiRoom-N4. Native limits remain
640 actions for DoorKey-8 and 40 for MultiRoom-N2. The added 200 episodes pass
stepwise replay verification of observations, query payloads, actions, and returns.
Exact added-task input examples are included with the source data.

\paragraph{Learning and scoring.}
MiniGrid DQN variants use an MLP, batch size 128, learning rate $2.5\times10^{-4}$,
discount $.99$, and one update per ten environment steps. Atari DQN variants use
a convolutional learner and one update per four decisions. Jev-PEX and Jev-BC use actor learners with MLPs for MiniGrid and CNNs for Atari; DQN is their common task baseline.
The DQN-based exploration and sampling variants retain the corresponding
base learner and change the designated intervention. Jev-BC uses
the imitation loss of Sec.~\ref{sec:jev-reference}. Jev-PEX fits two critics to
$r+\gamma(1-d)V(o')$, fits $V$ by expectile regression ($\tau=.7$) to the
minimum target-critic value of the recorded action, and uses fresh
composite-policy actions for its actor update. Target critics use Polyak
coefficient $.005$. The displayed action/transition bonuses use $\beta=.01$, except
FourRooms for both scores and DoorKey-8 for action novelty, which use $.1$;
state novelty uses $.1$. The coefficient and training-budget comparisons are
reported as task-specific configurations rather than a single universal setting.

To isolate which part of the learning procedure creates an observed gain, the
fixed-experience interventions retain this learner and change one source of
supervision at a time. The data-quality analysis trains on the first 10k chronological transitions from
each source with bonuses removed. Reward-label comparisons instead keep the
experience fixed and alter only its scores. Visitation bins are
$0,1,2,3$--$4,5$--$9,10$--$19,20$--$49,50+$, crossed with 2k collection windows;
the bin-mean control uses the recorded Jev means. Sampling analyses retain the
original rating history and change replay probabilities. All these diagnostic
fits use 899 gradient updates. Original state-score models were checked against
their stored 10k weights before the interventions. Source data for the bonus
collection diagnostic use $\beta=.1$. Spatial coverage counts newly reached
layout--position pairs in reconstructed trajectories; layout identity avoids
merging equal coordinates in different maps. The count-group control preserves
group means, while reward-class replay controls preserve total sampling weight
per reward class. Neither averaging operation preserves within-group variance.
For replay interventions, the recorded rating/refresh sequence is retained, so
the diagnostic changes sampling without re-querying Jev on the altered history.

\begin{figure}[!htbp]
 \centering\includegraphics[width=\linewidth]{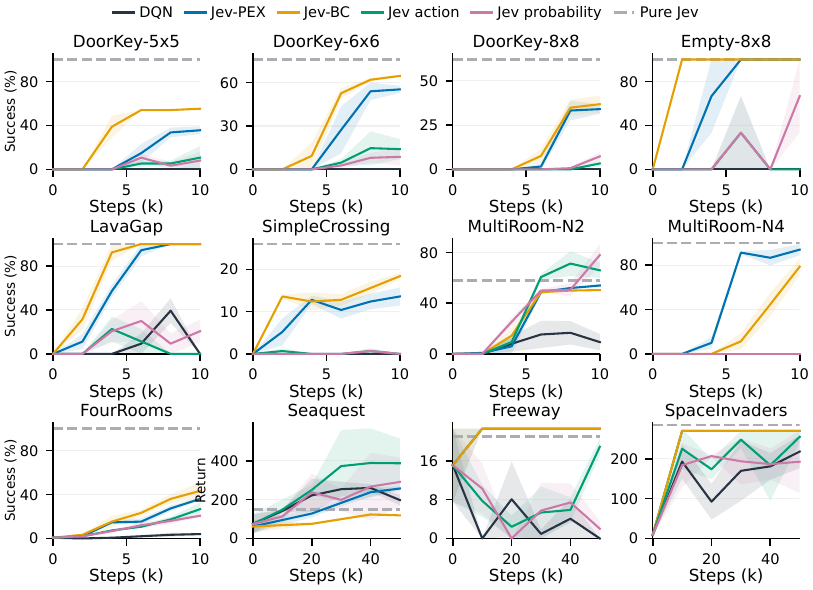}
 \caption{Reference-method learning curves, mean$\pm$SEM. The horizontal dashed
 line is the independently evaluated pure-Jev policy; it has no training-step dependence.}
 \label{fig:reference-curves}
\end{figure}

\begin{figure}[p]
 \centering\includegraphics[width=\linewidth]{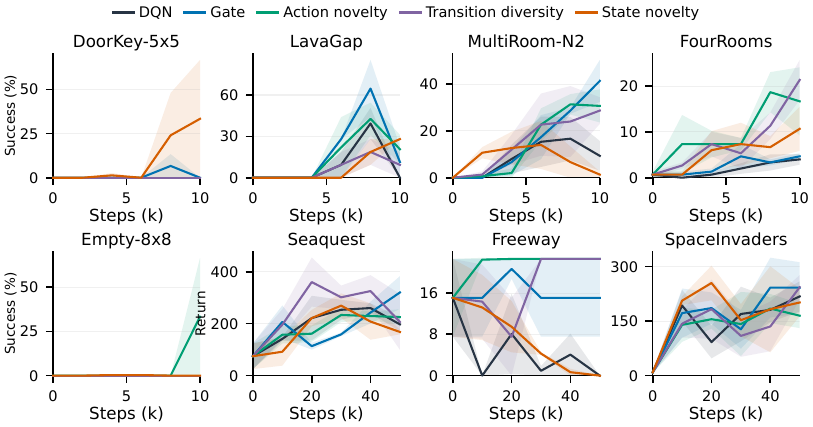}
 \caption{Exploration learning curves on the selected tasks, mean$\pm$SEM. Curves
 end at their last available common evaluation node.}
 \label{fig:exploration-curves}
\end{figure}

\begin{figure}[p]
 \centering\includegraphics[width=\linewidth]{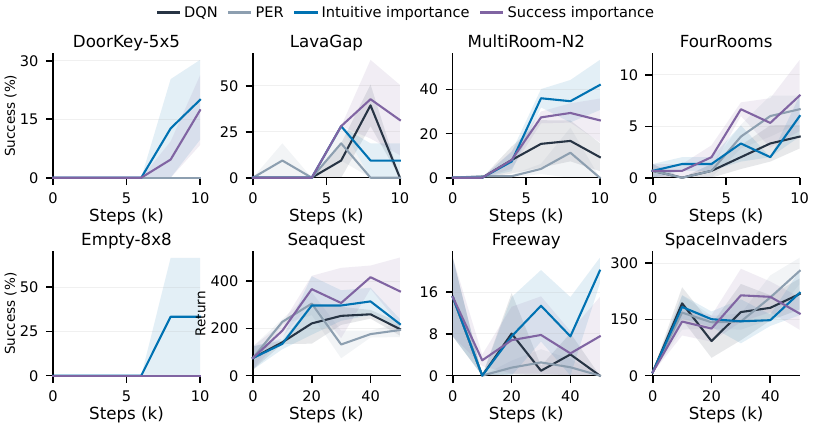}
 \caption{Replay-sampling learning curves on the selected tasks, mean$\pm$SEM.}
 \label{fig:sampling-curves}
\end{figure}

\paragraph{Learning curves.}
Figures~\ref{fig:reference-curves}--\ref{fig:sampling-curves} supplement the main
endpoint comparisons. All categories share the same DQN source. Seaquest Jev-PEX/Jev-BC
values at 50k are linearly interpolated within each training seed from 40k and
60k evaluations. Other missing endpoints are not filled by extrapolation. The
MiniGrid axes stop at 10k and Atari axes at 50k, matching the main figures.


\section{Analysis Experiments}
\label{sec:analysis-experiments}

\begin{figure}[!htbp]
  \centering
    \includegraphics[width=0.57\linewidth]{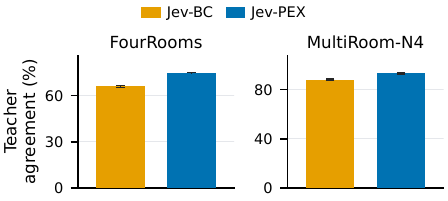}
    \caption{\textbf{Teacher-action absorption at 10k.}
    Agreement on fixed logged request states, averaged over three student seeds. Error bars show SEM.}
    \label{fig:teacher-uptake}
\end{figure}

We examine whether the students acquire Jev's action preferences by measuring agreement on fixed logged request states after 10k training steps. Figure~\ref{fig:teacher-uptake} shows substantial agreement for both Jev-BC and Jev-PEX, with Jev-PEX agreeing more often on both tasks. Nevertheless, both students remain below direct Jev control on FourRooms (Figure~\ref{fig:reference-bars}). The students therefore absorb considerable local action knowledge without necessarily matching the teacher's task performance. Action agreement and episode success measure different aspects of transfer: reproducing individual decisions does not guarantee successful execution over an entire episode.

\begin{figure}[!htbp]
  \centering
  \includegraphics[width=\linewidth]{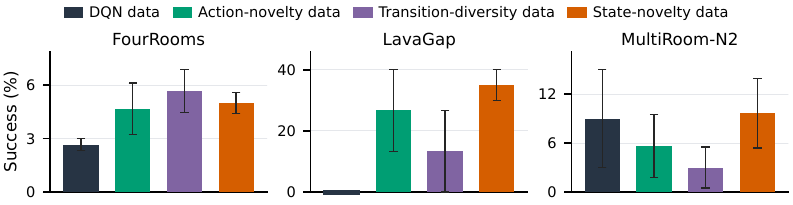}
  \caption{\textbf{Data quality separated from shaping.}
  All learners use only environment reward on their respective fixed 10k experience streams, with matched learner settings and sampling indices. Evaluation uses 100 environment seeds; bars show mean$\pm$SEM over three data/training seeds.}
  \label{fig:data-quality}
\end{figure}

To isolate the value of collected experience, we retrain identical learners on the first 10k transitions from DQN and each Jev-Bonus variant, removing all auxiliary rewards and matching learner settings and sampling indices. Figure~\ref{fig:data-quality} shows higher mean success with Jev-collected data on FourRooms and LavaGap, but mixed results on MultiRoom-N2. In particular, state-novelty data improves LavaGap learning despite having the same mean spatial coverage as DQN data. Thus, part of Jev's exploration benefit lies in collecting more useful experience, rather than simply visiting more locations or changing the training reward.

\begin{figure}[!htbp]
  \centering
  \includegraphics[width=\linewidth]{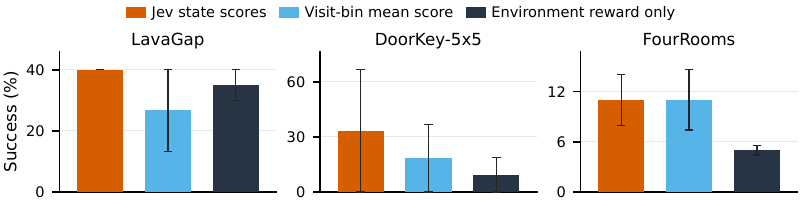}
  \caption{\textbf{State-dependent reward information on fixed experience.}
  The bin-mean control replaces scores with their average within 2k-step collection windows and visitation bins. Original and bin-mean scores have matched overall means but different fine-grained allocations. Bars show mean$\pm$SEM across three seeds; each fitted learner is evaluated on 100 episodes.}
  \label{fig:state-scores}
\end{figure}

We hold experience fixed and compare three reward assignments: the original Jev state bonuses, bonuses averaged within each visitation bin and 2k-step collection window, and environment reward alone. Averaging preserves each group's mean bonus but removes differences among observations within the group. Figure~\ref{fig:state-scores} shows that the original scores yield higher mean success than the averaged scores on LavaGap and DoorKey-5$\times$5, while both perform similarly on FourRooms. These results indicate that Jev's finer state-dependent distinctions can improve learning beyond the tested visitation summary, although the additional benefit is task-dependent.

\begin{figure}[!htbp]
  \centering
  \includegraphics[width=\linewidth]{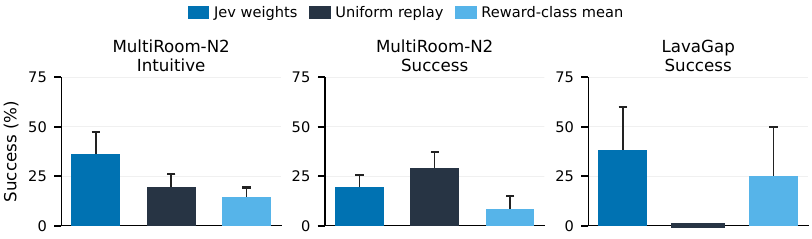}
  \caption{\textbf{Sampling-weight analysis on fixed experience.}
  Intuitive and success-oriented ratings are evaluated separately, using the same learner and original environment reward. Reward-class means preserve total preference for positive-reward transitions while removing within-class weight differences. Bars show mean$\pm$SEM across three seeds; each learner is evaluated on 100 episodes.}
  \label{fig:sampling-analysis}
\end{figure}

We fix each sampler's experience and rating history, then compare Jev weights with uniform replay and weights averaged separately within positive- and non-positive-reward classes. This control preserves each class's total sampling probability while treating its transitions equally. Figure~\ref{fig:sampling-analysis} shows that intuitive weights outperform both controls on MultiRoom-N2. Success-oriented weights, evaluated on separately collected data, outperform the reward-class control but underperform uniform replay on the same task. On LavaGap, success-oriented weights yield the highest mean success, although variability across seeds is large. Jev can therefore improve replay by distinguishing useful transitions beyond their immediate rewards, but following its ratings is not consistently better than sampling uniformly.
\section{Jev for Subgoal Guidance}
\label{app:subgoal}


This appendix reports a fourth channel, subgoal selection, in which the same model is
asked to guide behavior toward an event that requires several actions to complete. We
report it here rather than in the main text because it did not yield gains under the
shared protocol, and because its diagnostics bear on a different requirement than the
three channels in Sec.~\ref{sec:method}.

\subsection{Channel}
\label{app:subgoal-channel}

The channels in Sec.~\ref{sec:method} attach judgments to actions, states, or transitions.
Subgoal selection asks whether the same model can also guide behavior toward an
event that requires several actions to complete. For MiniGrid, the adapter constructs a finite candidate
set $\mathcal G_t$ from observed objects and connectivity: for example, obtaining
a key, opening a door, traversing a passage, or reaching the goal. The query
supplies these candidates, observed-map information, prerequisites, and previous
attempts. Jev selects
\begin{equation}
 g_t=\arg\max_{g\in\mathcal G_t}f_{u_{\mathrm{goal}}}(x_t,H_t)_g.
 \label{eq:jev-goal}
\end{equation}
The selected milestone determines an auxiliary completion reward:
\begin{equation}
 r'_t=r_t+\alpha\,\mathbf 1\{\text{first eligible completion of }g_t\},
 \qquad\alpha=.025.
 \label{eq:jev-goal-reward}
\end{equation}
The cumulative bonus is capped at $.1$ per episode. An environment-side check
determines completion; the selector is queried again after completion, timeout,
or prolonged lack of progress. Repeating a completed milestone earns no further
reward, although an interaction can remain a selectable prerequisite when needed
again, as after dropping a key.

In this primary MiniGrid comparison, the student still selects primitive actions
from its ordinary observation and does not receive $g_t$ as an additional input.
Thus, the intervention is milestone-based reward guidance rather than a learned
hierarchy of goal-conditioned options. Uniform and rule-based selectors share
the same candidate set and reward mechanism with Jev. Additional diagnostics
change the controller input or provide denser progress rewards. This separation
lets us examine two requirements for effective guidance: a goal must be useful
for the task, and its completion must deliver a learnable signal to the current
controller.

\subsection{Results}
\label{app:subgoal-results}

The channels in Sec.~\ref{sec:method} guide choices whose consequences enter
learning through an action or transition. A selected subgoal demands more: the
controller must produce a sequence that reaches it before the auxiliary reward can
be earned. We first test whether meaningful milestones alone provide effective guidance.
In DoorKey-8$\times$8 and FourRooms, DQN, uniform goal selection, Jev goal
selection, and the rule selector all finish at zero success after 20k steps in
the single-seed diagnostic with corrected candidate eligibility and one-time
rewards. FourRooms briefly reaches 10\% with Jev at 8k, but the gain does not
persist. Rule-based variants with progress rewards or explicit goal inputs also
finish at zero. These controls locate a difficulty shared by the tested
milestone-reward channels, rather than a failure unique to Jev's ranking.

The sparse milestone reward may be difficult to obtain from a controller
learning from scratch. To examine goal selection after shared low-level
pretraining, a separate Atari diagnostic uses 50k steps of that pretraining and
50k subsequent student steps. Jev probabilities, weighted toward less frequently
completed goals, yield 382.7 on Seaquest versus 345.3 for uniform goals, but 17.2
on Freeway versus 19.7. Thus the 100k-interaction protocol also gives mixed
evidence for choosing goals semantically, even with a pretrained controller.

Mixed returns alone do not show whether the chosen goals produce usable reward
events. Historical Montezuma's Revenge runs let us inspect this part of the
mechanism directly.
The controller is image-only PPO+RND \citep{schulman2017ppo,burda2019rnd}; Jev
selects landmarks whose completion provides auxiliary reward. In the common
12M-decision training prefix, pooled across three seeds, 77.1\% of goal selections
are allocated to obtaining the key. That goal is selected 481,153 times and
completed only twice. Overall selected-goal completion is 0.36\%, compared with
1.98\% for uniform selection at the same reward coefficient. Selecting a valuable
eventual objective repeatedly therefore produces very little realized
supervision. These are recorded goal-selection events, including reuse of cached
Jev answers, rather than distinct API calls. The historical query supplies the
visual scene and task rules, but not measured success rates for the current
controller. Its preference for the key therefore captures task importance without
explicit evidence of present attainability.

The missing property is \emph{effective temporal guidance}: a useful selector
must connect task prerequisites to milestones the current controller can achieve
within its execution horizon. Our subgoal channel does not obtain that property
reliably from Jev. The rule controls and completion audit further show why
correctly naming the key or door is insufficient evidence of long-horizon
planning ability. Across the tested systems, local action and experience
judgments are easier to turn into learning gains than milestone preferences
whose value depends on an executable sequence.

\clearpage
\raggedbottom
\section{Structured Input Examples}
\label{app:complete-inputs}
The cards below give one short example for each environment and input setting.
Each card first shows a concrete scene, then examples from each method's own
logged request. Different rows may therefore describe different moments.
Current, before/after, and reached denote the actual scene fields included in
that query, not only visit counts. Long object lists and histories are abbreviated;
field paths are shortened for readability.
Tables~\ref{tab:exploration-prompts} and~\ref{tab:other-prompts} separate the
shared conventions from each method's question, evidence and output. The cards
then give environment-specific values. Table~\ref{tab:query-memory} specifies
which recent states or transitions are supplied, their maximum lengths, and
whether they survive an episode reset. Matching fields in two rows do not imply
matching questions: replay importance is one explicit example.

\newtcolorbox{inputexample}[2][]{enhanced,colback=blue!2!white,
 colframe=blue!35!black,colbacktitle=blue!8!white,coltitle=black,
 boxrule=.45pt,arc=1.2mm,left=7pt,right=7pt,top=5pt,bottom=5pt,
 fonttitle=\small\bfseries,fontupper=\small,title={#2},
 before skip=9pt,after skip=9pt,#1}

\newcommand{\querydiff}[1]{\textcolor{blue!55!black}{\textbf{#1}}}
\begin{inputexample}{Shared request structure and answer scale}
Each request has a \texttt{state} and a \texttt{questions} object. A question
contains \texttt{instructions} and named answer \texttt{criteria}. The environment
cards below supply example state fields; the tables here show what Jev is asked
to do with them. Quoted sentences are excerpts from recorded MiniGrid instructions; Atari queries
use the corresponding game facts and action meanings.
\querydiff{Bold blue text} marks the operative difference between queries.

The three novelty bonuses and both importance scores share the ordered choices
\texttt{zero, low, medium, high, very\_high, unknown}, mapped to
$0,.25,.5,.75,1,0$. Each question returns its own distribution over these levels;
its expected numerical value is the score. A rating is not a probability of
selecting a primitive action.
\end{inputexample}

\begin{table}[H]
\begin{inputexample}{Exploration: what question changes?}
\setlength{\tabcolsep}{4pt}\renewcommand{\arraystretch}{1.12}
\begin{tabular}{@{}p{.155\linewidth}p{.29\linewidth}p{.50\linewidth}@{}}
\textbf{Method} & \textbf{Evidence and output} & \textbf{Instruction excerpt and distinction}\\
\midrule
\textbf{Gate} & Before acting: current scene, action effects, experience history,
and \querydiff{student's greedy action / Q-values}. Output: GREEDY or RANDOM. &
``Choose whether to \querydiff{exploit this learner's CURRENT greedy action or
explore with a uniformly random legal action NOW}.''
``You do not choose the primitive action.'' The answer selects a control branch,
not a teacher action or reward.\\[6pt]
\textbf{Action novelty} & Before acting: current scene, candidate action effects,
per-action counts and past outcomes. Output: \querydiff{one rating per action}. &
``Independently rate the \querydiff{EXPECTED novelty} of the controllable situation
or action-effect information that would result from this particular action after
one native MiniGrid action.'' The score predicts an outcome; only the executed
action's score becomes a bonus.\\[6pt]
\textbf{Transition diversity} & After acting: \querydiff{before state, executed
action, actual after state}, observed reward and lifetime experience counts.
Output: one rating. &
``Rate the \querydiff{ACTUAL contribution of this observed transition} to the
diversity of controllable experiences in THIS training run.''
``Judge what actually happened, not what this action might have achieved.''
The realized transition is the object being scored.\\[6pt]
\textbf{State novelty} & After acting: \querydiff{reached observation and
state-visit history only}. Output: one rating. No action, reward, Q-value or
before/after pair is supplied. &
``Rate the novelty of the \querydiff{CANDIDATE OBSERVED STATE} relative to states
previously observed in THIS training run.''
``Only judge the state, not an action, transition, expected outcome, reward
desirability or task success.''\\
\end{tabular}

\smallskip
\textbf{Shared novelty instruction.} Judge unfamiliar controllable experience
relative to this run, rather than desirability or action quality. Familiar loops,
passive visual changes, deaths and resets do not earn novelty. Counts persist
across episodes; limited recent history is not proof that an outcome is unseen.
The action-novelty time horizon uses one native action in MiniGrid and the
corresponding decision transition in Atari.
\end{inputexample}
\caption{Exploration query comparison. The queries derive from the same environment observations, but
 each method exposes only the evidence listed in its row. The three bonuses
 share a rating scale, not a common question.}
\label{tab:exploration-prompts}
\end{table}

\begin{table}[H]
\begin{inputexample}{Reference, replay sampling, and subgoals}
\textbf{Shared replay instruction.} Both importance variants receive the same
\texttt{transition} and \texttt{buffer\_evidence}. Their common prefix is:
``Assess this OBSERVED transition for replay learning. You do not select an action
or modify rewards.'' The instruction below changes the criterion, not the input
fields, rating options, or conversion from score to replay priority.

\setlength{\tabcolsep}{4pt}\renewcommand{\arraystretch}{1.12}
\begin{tabular}{@{}p{.23\linewidth}p{.73\linewidth}@{}}
\textbf{Setting} & \textbf{Instruction excerpt and use of the answer}\\
\midrule
\textbf{Action reference} &
``Which single action should the agent \querydiff{execute NEXT to reach the goal}?''
The state describes the scene, task rules and immediate action effects; options
are legal actions. Jev-BC, Jev-PEX, Jev action, and Jev probability sampling use
this action-preference question. Their downstream use of the answer differs.\\[6pt]
\textbf{Intuitive importance} &
``How important is this experience to replay for learning this task?
\querydiff{Use your own judgment of importance}.'' The score directly sets
replay priority; no particular type of importance is specified.\\[6pt]
\textbf{Success importance} &
``Judge ONLY its contribution to learning \querydiff{HOW TO ACHIEVE THE FINAL TASK
OBJECTIVE}.'' ``Prioritize evidenced prerequisites, bottleneck interactions,
meaningful progress, successful completion, and informative contrasts that help
explain how completion/high task return becomes possible.'' Zero immediate
reward can still be important.\\[6pt]
\textbf{Subgoal selection} &
``Select ONE candidate subgoal that is \querydiff{useful for eventually completing
the mission and feasible for the current learner to explore next}.''
``You choose a target, not a primitive action or a route.'' The input adds
observed candidates, prerequisites and previous attempts. Completing the selected
milestone triggers the auxiliary reward.\\
\end{tabular}
\end{inputexample}
\caption{Other query settings. In particular, identical replay state examples
 do not imply identical questions: the two importance variants differ in the
 highlighted criterion.}
\label{tab:other-prompts}
\end{table}

\begin{inputexample}{Shared action and coordinate conventions}
MiniGrid uses \texttt{turn\_left, turn\_right, move\_forward, pick\_up, drop,
toggle\_door, done}; their physical meanings accompany action queries. Relative
object coordinates are \texttt{(cells\_forward, cells\_right)} in the agent's view.
Atari coordinates are RGB screen pixels, with $x$ increasing right and $y$
increasing down; legal action sets appear in each environment card.
DQN, PER and rule/uniform-goal controls make no Jev query. Reward-coefficient and
update-frequency variants reuse the same query. The cards show abbreviated
state values; use the method's table row above to identify its instruction.
\end{inputexample}

\begin{table}[H]
\begin{inputexample}{Memory supplied to Jev: contents, length, and reset scope}
\setlength{\tabcolsep}{4pt}\renewcommand{\arraystretch}{1.12}
\begin{tabular}{@{}p{.19\linewidth}p{.44\linewidth}p{.32\linewidth}@{}}
\textbf{Query} & \textbf{Records included in the prompt} & \textbf{Counts and reset scope}\\
\midrule
Gate / action novelty & Up to \querydiff{16 recent transitions}: coarse before
situation, action, coarse after situation, reward and boundary flag. For each
candidate action, up to \querydiff{3 most frequent observed outcomes}, with counts.
A separate 32-transition window contributes a distinct-situation count. &
Visits, action attempts and outcome counts accumulate over the training run.
The recent trace and 32-transition window clear at a new complete game/episode;
lifetime counts remain.\\[6pt]
Transition diversity & The same recent-16 trace and per-action top-3 outcomes,
plus the actual candidate's before/after scene and action. &
Adds lifetime counts of matching transitions and visits to the reached situation.
The candidate is scored \querydiff{before} updating its transition/action/outcome
counts.\\[6pt]
State novelty & Up to \querydiff{16 recent states} and \querydiff{3 most-visited
states}, each with its lifetime visit count. MiniGrid records contain the
\querydiff{full local $7\times7$ view and compass}; Atari history records contain
\querydiff{coarse player/resource situations}. No actions or rewards are included. &
Both recent history and lifetime counts persist across resets. The candidate's
previous-visit count is read before adding it. Recent entries may repeat: they
are observations, not a list of unique states.\\[6pt]
Replay importance & Up to \querydiff{12 recent observed transitions}, with
before/after situations, action, reward and termination flags; matching counts,
outcome counts and the candidate's replay count. &
Matching counts describe the \querydiff{current replay buffer}; eviction removes
counts and the candidate is excluded. The recent trace persists across episodes
and is a chronological record, not a fresh sample of the buffer.\\
\end{tabular}
\smallskip
The bounded lists are actual evidence in the query, not a recurrent hidden state
of Jev. Each decision is made from a fresh request. ``Not present in the recent
list'' does not mean ``never visited'': the accompanying counts summarize a
larger experience history. Atari coarse states bin controlled-player position
in 16-pixel cells; Seaquest additionally includes oxygen, carried-diver icons and
surface status. MiniGrid identity uses the local observation and compass,
excluding elapsed time. These are observation identities, not hidden full-map
coordinates.
\end{inputexample}
\caption{The memory interface differs across exploration and sampling methods.
Short state/transition lists and cumulative counts are distinct inputs.}
\label{tab:query-memory}
\end{table}

\begin{inputexample}{Navigation memory is a different input}
FourRooms and MultiRoom-N4 action queries additionally use within-episode
observed-map memory: cells reconstructed from past local views, the relative
agent pose and its visit count, and the last eight action/movement records.
Observed candidate targets and route costs are derived from that map. This
memory clears on reset. It is distinct from the recent-state lists and lifetime
experience counts used by the exploration bonuses. The navigation ablation
removes candidate routes and their progress/detour evidence while retaining the
observational memory specified in the ablation interface.
\end{inputexample}

\clearpage

\begin{inputexample}{DoorKey-5x5}

\textbf{Task information.} Collect the matching key, unlock the door, then enter the goal.

\textbf{Action choices.} The seven MiniGrid actions above.

\textbf{Concrete scene (action-query example).} {\footnotesize\ttfamily \{"ahead": "wall", "left": "empty", "right": "empty", "carrying": "yellow\_\allowbreak key", "visible": [["yellow\_\allowbreak locked\_\allowbreak door", 1, -1]]\}}\par\smallskip
\begin{minipage}[t]{.25\linewidth}\footnotesize\ttfamily
\setlength{\tabcolsep}{3pt}\begin{tabular}{@{}rrrrrrr@{}}
? & ? & ? & ? & ? & ? & ?\\
? & ? & ? & ? & ? & ? & ?\\
? & ? & ? & ? & ? & ? & ?\\
? & ? & ? & ? & ? & ? & ?\\
? & ? & ? & ? & ? & ? & ?\\
? & W & D & W & W & W & ?\\
? & W & . & K & . & W & ?\\
\end{tabular}\end{minipage}\hfill\begin{minipage}[t]{.71\linewidth}\footnotesize
The complete native $7\times7$ view for this example. Rows run top to bottom; the agent occupies row 6, column 3 (zero-based), facing row 5. The agent cell encodes its carried object. \texttt{?}: unseen; \texttt{.}: empty; \texttt{W}: wall; \texttt{K}: key; \texttt{D}: door; \texttt{G}: goal; \texttt{L}: lava. Colors and door status are retained in the actual request.
\end{minipage}\par\smallskip

\end{inputexample}
\begin{inputexample}{DoorKey-5x5 --- method-specific state examples}

\setlength{\tabcolsep}{3pt}\renewcommand{\arraystretch}{1.05}
\begin{tabular}{@{}p{.235\linewidth}p{.735\linewidth}@{}}

\textbf{Action reference} & {\footnotesize\ttfamily \{"current": \{"ahead": "wall", "carrying": "yellow\_\allowbreak key", "visible": [["yellow\_\allowbreak locked\_\allowbreak door", 1, -1]]\}\}}\\[3pt]

\textbf{Gate} & {\footnotesize\ttfamily \{"current": \{"ahead": "wall", "carrying": "nothing", "visible": [["yellow\_\allowbreak locked\_\allowbreak door", 1, 1], ["yellow\_\allowbreak key", 0, 2]]\}, "greedy\_\allowbreak action": "done", "learner\_\allowbreak updates": 0, "current\_\allowbreak situation\_\allowbreak visits": 2\}}\\[3pt]

\textbf{Action novelty} & {\footnotesize\ttfamily \{"current": \{"ahead": "wall", "carrying": "nothing", "visible": [["yellow\_\allowbreak locked\_\allowbreak door", 1, 1], ["yellow\_\allowbreak key", 0, 2]]\}, "action\_\allowbreak to\_\allowbreak rate": "turn\_\allowbreak left", "attempts\_\allowbreak in\_\allowbreak current\_\allowbreak situation": 0, "current\_\allowbreak situation\_\allowbreak visits": 2\}}\\[3pt]

\textbf{Transition diversity} & {\footnotesize\ttfamily \{"before": \{"ahead": "wall", "carrying": "nothing", "visible": [["yellow\_\allowbreak locked\_\allowbreak door", 1, 1], ["yellow\_\allowbreak key", 0, 2]]\}, "after": \{"ahead": "wall", "carrying": "nothing", "visible": []\}, "action": "turn\_\allowbreak left", "reward": 0.0, "previous\_\allowbreak transition\_\allowbreak count": 0, "after\_\allowbreak previous\_\allowbreak visits": 7\}}\\[3pt]

\textbf{State novelty} & {\footnotesize\ttfamily \{"reached": \{"compass": "south", "row5": ["unseen", "unseen", "wall", "wall", "wall", "unseen", "unseen"]\}, "candidate\_\allowbreak previous\_\allowbreak visits": 8, "recent\_\allowbreak state\_\allowbreak example": \{"state": \{"compass": "east", "row5": ["wall", "yellow\_\allowbreak locked\_\allowbreak door", "wall", "wall", "wall", "unseen", "unseen"]\}, "lifetime\_\allowbreak visits": 1\}\}}\\[3pt]

\textbf{Replay input (shared)} & {\footnotesize\ttfamily \{"before": \{"ahead": "wall", "carrying": "nothing", "visible": []\}, "after": \{"ahead": "wall", "carrying": "nothing", "visible": []\}, "action": "pick\_\allowbreak up", "reward": 0.0, "matching\_\allowbreak transitions": 0, "goal\_\allowbreak entry": false\}}\\[3pt]

\end{tabular}
\end{inputexample}

\clearpage

\begin{inputexample}{DoorKey-6x6}

\textbf{Task information.} Collect the matching key, unlock the door, then enter the goal.

\textbf{Action choices.} The seven MiniGrid actions above.

\textbf{Concrete scene (action-query example).} {\footnotesize\ttfamily \{"ahead": "yellow\_\allowbreak key", "left": "wall", "right": "wall", "carrying": "nothing", "visible": [["yellow\_\allowbreak key", 1, 0], ["goal", 1, 3]]\}}\par\smallskip
\begin{minipage}[t]{.25\linewidth}\footnotesize\ttfamily
\setlength{\tabcolsep}{3pt}\begin{tabular}{@{}rrrrrrr@{}}
? & ? & ? & ? & ? & ? & ?\\
? & ? & ? & ? & ? & ? & ?\\
? & ? & ? & ? & ? & ? & ?\\
? & ? & ? & ? & ? & ? & ?\\
? & ? & W & W & W & W & W\\
? & ? & W & K & . & . & G\\
? & ? & W & . & W & ? & ?\\
\end{tabular}\end{minipage}\hfill\begin{minipage}[t]{.71\linewidth}\footnotesize
The complete native $7\times7$ view for this example. Rows run top to bottom; the agent occupies row 6, column 3 (zero-based), facing row 5. The agent cell encodes its carried object. \texttt{?}: unseen; \texttt{.}: empty; \texttt{W}: wall; \texttt{K}: key; \texttt{D}: door; \texttt{G}: goal; \texttt{L}: lava. Colors and door status are retained in the actual request.
\end{minipage}\par\smallskip

\end{inputexample}
\begin{inputexample}{DoorKey-6x6 --- method-specific state examples}

\setlength{\tabcolsep}{3pt}\renewcommand{\arraystretch}{1.05}
\begin{tabular}{@{}p{.235\linewidth}p{.735\linewidth}@{}}

\textbf{Jev-BC / Jev-PEX} & {\footnotesize\ttfamily \{"current": \{"ahead": "yellow\_\allowbreak key", "carrying": "nothing", "visible": [["yellow\_\allowbreak key", 1, 0], ["goal", 1, 3]]\}\}}\\[3pt]

\textbf{Jev action / probability} & {\footnotesize\ttfamily \{"current": \{"ahead": "empty", "carrying": "nothing", "visible": [["yellow\_\allowbreak key", 1, -1], ["yellow\_\allowbreak locked\_\allowbreak door", 0, -2]]\}\}}\\[3pt]

\textbf{Gate} & {\footnotesize\ttfamily \{"current": \{"ahead": "wall", "carrying": "nothing", "visible": [["yellow\_\allowbreak key", 0, 3]]\}, "greedy\_\allowbreak action": "done", "learner\_\allowbreak updates": 0, "current\_\allowbreak situation\_\allowbreak visits": 7\}}\\[3pt]

\textbf{Action novelty} & {\footnotesize\ttfamily \{"current": \{"ahead": "wall", "carrying": "nothing", "visible": [["yellow\_\allowbreak key", 0, 3]]\}, "action\_\allowbreak to\_\allowbreak rate": "turn\_\allowbreak left", "attempts\_\allowbreak in\_\allowbreak current\_\allowbreak situation": 1, "current\_\allowbreak situation\_\allowbreak visits": 7\}}\\[3pt]

\textbf{Transition diversity} & {\footnotesize\ttfamily \{"before": \{"ahead": "wall", "carrying": "nothing", "visible": [["yellow\_\allowbreak key", 0, 3]]\}, "after": \{"ahead": "wall", "carrying": "nothing", "visible": []\}, "action": "turn\_\allowbreak left", "reward": 0.0, "previous\_\allowbreak transition\_\allowbreak count": 1, "after\_\allowbreak previous\_\allowbreak visits": 8\}}\\[3pt]

\textbf{Replay input (shared)} & {\footnotesize\ttfamily \{"before": \{"ahead": "wall", "carrying": "nothing", "visible": []\}, "after": \{"ahead": "wall", "carrying": "nothing", "visible": []\}, "action": "pick\_\allowbreak up", "reward": 0.0, "matching\_\allowbreak transitions": 0, "goal\_\allowbreak entry": false\}}\\[3pt]

\end{tabular}
\end{inputexample}

\clearpage

\begin{inputexample}{DoorKey-8x8}

\textbf{Task information.} Collect the matching key, unlock the door, then enter the goal.

\textbf{Action choices.} The seven MiniGrid actions above.

\textbf{Concrete scene (action-query example).} {\footnotesize\ttfamily \{"ahead": "yellow\_\allowbreak key", "left": "wall", "right": "empty", "carrying": "nothing", "visible": [["yellow\_\allowbreak key", 1, 0]]\}}\par\smallskip
\begin{minipage}[t]{.25\linewidth}\footnotesize\ttfamily
\setlength{\tabcolsep}{3pt}\begin{tabular}{@{}rrrrrrr@{}}
? & ? & ? & ? & ? & ? & ?\\
? & ? & ? & ? & ? & ? & ?\\
? & ? & W & W & W & W & W\\
? & ? & W & . & . & . & .\\
? & ? & W & . & . & . & .\\
? & ? & W & K & . & . & .\\
? & ? & W & . & . & . & .\\
\end{tabular}\end{minipage}\hfill\begin{minipage}[t]{.71\linewidth}\footnotesize
The complete native $7\times7$ view for this example. Rows run top to bottom; the agent occupies row 6, column 3 (zero-based), facing row 5. The agent cell encodes its carried object. \texttt{?}: unseen; \texttt{.}: empty; \texttt{W}: wall; \texttt{K}: key; \texttt{D}: door; \texttt{G}: goal; \texttt{L}: lava. Colors and door status are retained in the actual request.
\end{minipage}\par\smallskip

\end{inputexample}
\begin{inputexample}{DoorKey-8x8 --- method-specific state examples}

\setlength{\tabcolsep}{3pt}\renewcommand{\arraystretch}{1.05}
\begin{tabular}{@{}p{.235\linewidth}p{.735\linewidth}@{}}

\textbf{Jev-BC / Jev-PEX} & {\footnotesize\ttfamily \{"current": \{"ahead": "yellow\_\allowbreak key", "carrying": "nothing", "visible": [["yellow\_\allowbreak key", 1, 0]]\}\}}\\[3pt]

\textbf{Jev action / probability} & {\footnotesize\ttfamily \{"current": \{"ahead": "empty", "carrying": "nothing", "visible": [["yellow\_\allowbreak key", 1, -1]]\}\}}\\[3pt]

\textbf{Gate} & {\footnotesize\ttfamily \{"current": \{"ahead": "wall", "carrying": "nothing", "visible": []\}, "greedy\_\allowbreak action": "done", "learner\_\allowbreak updates": 0, "current\_\allowbreak situation\_\allowbreak visits": 7\}}\\[3pt]

\textbf{Action novelty} & {\footnotesize\ttfamily \{"current": \{"ahead": "empty", "carrying": "nothing", "visible": [["yellow\_\allowbreak locked\_\allowbreak door", 2, 2]]\}, "action\_\allowbreak to\_\allowbreak rate": "turn\_\allowbreak left", "attempts\_\allowbreak in\_\allowbreak current\_\allowbreak situation": 0, "current\_\allowbreak situation\_\allowbreak visits": 1\}}\\[3pt]

\textbf{Transition diversity} & {\footnotesize\ttfamily \{"before": \{"ahead": "wall", "carrying": "nothing", "visible": []\}, "after": \{"ahead": "wall", "carrying": "nothing", "visible": []\}, "action": "turn\_\allowbreak left", "reward": 0.0, "previous\_\allowbreak transition\_\allowbreak count": 1, "after\_\allowbreak previous\_\allowbreak visits": 8\}}\\[3pt]

\textbf{Replay input (shared)} & {\footnotesize\ttfamily \{"before": \{"ahead": "wall", "carrying": "nothing", "visible": []\}, "after": \{"ahead": "wall", "carrying": "nothing", "visible": []\}, "action": "pick\_\allowbreak up", "reward": 0.0, "matching\_\allowbreak transitions": 0, "goal\_\allowbreak entry": false\}}\\[3pt]

\textbf{Subgoal} & {\footnotesize\ttfamily \{"current": \{"ahead": "empty", "left": "empty", "right": "empty", "carrying": "nothing", "visible": [["yellow\_\allowbreak key", 1, -1]]\}, "kind": "collect\_\allowbreak key", "object": "yellow\_\allowbreak key", "reward\_\allowbreak eligible": true, "completed\_\allowbreak before": false\}}\\[3pt]

\end{tabular}
\end{inputexample}

\clearpage

\begin{inputexample}{Empty-8x8}

\textbf{Task information.} Navigate to the green goal using the native partial view.

\textbf{Action choices.} The seven MiniGrid actions above.

\textbf{Concrete scene (action-query example).} {\footnotesize\ttfamily \{"ahead": "empty", "left": "wall", "right": "empty", "carrying": "nothing", "visible": []\}}\par\smallskip
\begin{minipage}[t]{.25\linewidth}\footnotesize\ttfamily
\setlength{\tabcolsep}{3pt}\begin{tabular}{@{}rrrrrrr@{}}
W & W & W & W & W & W & W\\
W & W & W & W & W & W & W\\
W & W & W & W & W & W & W\\
W & W & W & W & W & W & W\\
W & W & W & . & . & . & .\\
W & W & W & . & . & . & .\\
W & W & W & . & . & . & .\\
\end{tabular}\end{minipage}\hfill\begin{minipage}[t]{.71\linewidth}\footnotesize
The complete native $7\times7$ view for this example. Rows run top to bottom; the agent occupies row 6, column 3 (zero-based), facing row 5. The agent cell encodes its carried object. \texttt{?}: unseen; \texttt{.}: empty; \texttt{W}: wall; \texttt{K}: key; \texttt{D}: door; \texttt{G}: goal; \texttt{L}: lava. Colors and door status are retained in the actual request.
\end{minipage}\par\smallskip

\end{inputexample}
\begin{inputexample}{Empty-8x8 --- method-specific state examples}

\setlength{\tabcolsep}{3pt}\renewcommand{\arraystretch}{1.05}
\begin{tabular}{@{}p{.235\linewidth}p{.735\linewidth}@{}}

\textbf{Jev-BC / Jev-PEX} & {\footnotesize\ttfamily \{"current": \{"ahead": "empty", "carrying": "nothing", "visible": []\}\}}\\[3pt]

\textbf{Jev action / probability} & {\footnotesize\ttfamily \{"current": \{"ahead": "empty", "carrying": "nothing", "visible": [["goal", 4, 0]]\}\}}\\[3pt]

\textbf{Gate} & {\footnotesize\ttfamily \{"current": \{"ahead": "wall", "carrying": "nothing", "visible": []\}, "greedy\_\allowbreak action": "done", "learner\_\allowbreak updates": 0, "current\_\allowbreak situation\_\allowbreak visits": 7\}}\\[3pt]

\textbf{Action novelty} & {\footnotesize\ttfamily \{"current": \{"ahead": "wall", "carrying": "nothing", "visible": []\}, "action\_\allowbreak to\_\allowbreak rate": "turn\_\allowbreak left", "attempts\_\allowbreak in\_\allowbreak current\_\allowbreak situation": 1, "current\_\allowbreak situation\_\allowbreak visits": 7\}}\\[3pt]

\textbf{Transition diversity} & {\footnotesize\ttfamily \{"before": \{"ahead": "wall", "carrying": "nothing", "visible": []\}, "after": \{"ahead": "wall", "carrying": "nothing", "visible": []\}, "action": "turn\_\allowbreak left", "reward": 0.0, "previous\_\allowbreak transition\_\allowbreak count": 1, "after\_\allowbreak previous\_\allowbreak visits": 8\}}\\[3pt]

\textbf{Replay input (shared)} & {\footnotesize\ttfamily \{"before": \{"ahead": "wall", "carrying": "nothing", "visible": []\}, "after": \{"ahead": "wall", "carrying": "nothing", "visible": []\}, "action": "pick\_\allowbreak up", "reward": 0.0, "matching\_\allowbreak transitions": 0, "goal\_\allowbreak entry": false\}}\\[3pt]

\end{tabular}
\end{inputexample}

\clearpage

\begin{inputexample}{LavaGap}

\textbf{Task information.} Find the safe gap; stepping into lava ends the episode.

\textbf{Action choices.} The seven MiniGrid actions above.

\textbf{Concrete scene (action-query example).} {\footnotesize\ttfamily \{"ahead": "empty", "left": "lava", "right": "wall", "carrying": "nothing", "visible": [["goal", 2, -2], ["lava", 2, -1]]\}}\par\smallskip
\begin{minipage}[t]{.25\linewidth}\footnotesize\ttfamily
\setlength{\tabcolsep}{3pt}\begin{tabular}{@{}rrrrrrr@{}}
? & ? & ? & ? & ? & ? & ?\\
? & ? & ? & ? & ? & ? & ?\\
? & ? & ? & ? & ? & ? & ?\\
W & W & W & W & W & ? & ?\\
W & G & L & . & W & ? & ?\\
W & . & . & . & W & ? & ?\\
W & . & L & . & W & ? & ?\\
\end{tabular}\end{minipage}\hfill\begin{minipage}[t]{.71\linewidth}\footnotesize
The complete native $7\times7$ view for this example. Rows run top to bottom; the agent occupies row 6, column 3 (zero-based), facing row 5. The agent cell encodes its carried object. \texttt{?}: unseen; \texttt{.}: empty; \texttt{W}: wall; \texttt{K}: key; \texttt{D}: door; \texttt{G}: goal; \texttt{L}: lava. Colors and door status are retained in the actual request.
\end{minipage}\par\smallskip

\end{inputexample}
\begin{inputexample}{LavaGap --- method-specific state examples}

\setlength{\tabcolsep}{3pt}\renewcommand{\arraystretch}{1.05}
\begin{tabular}{@{}p{.235\linewidth}p{.735\linewidth}@{}}

\textbf{Jev-BC / Jev-PEX} & {\footnotesize\ttfamily \{"current": \{"ahead": "empty", "carrying": "nothing", "visible": [["goal", 2, -2], ["lava", 2, -1]]\}\}}\\[3pt]

\textbf{Jev action / probability} & {\footnotesize\ttfamily \{"current": \{"ahead": "empty", "carrying": "nothing", "visible": [["goal", 2, 2], ["lava", 1, 1]]\}\}}\\[3pt]

\textbf{Gate} & {\footnotesize\ttfamily \{"current": \{"ahead": "wall", "carrying": "nothing", "visible": [["lava", 0, 1]]\}, "greedy\_\allowbreak action": "done", "learner\_\allowbreak updates": 0, "current\_\allowbreak situation\_\allowbreak visits": 7\}}\\[3pt]

\textbf{Action novelty} & {\footnotesize\ttfamily \{"current": \{"ahead": "wall", "carrying": "nothing", "visible": [["lava", 0, 1]]\}, "action\_\allowbreak to\_\allowbreak rate": "turn\_\allowbreak left", "attempts\_\allowbreak in\_\allowbreak current\_\allowbreak situation": 1, "current\_\allowbreak situation\_\allowbreak visits": 7\}}\\[3pt]

\textbf{Transition diversity} & {\footnotesize\ttfamily \{"before": \{"ahead": "wall", "carrying": "nothing", "visible": [["lava", 0, 1]]\}, "after": \{"ahead": "wall", "carrying": "nothing", "visible": []\}, "action": "turn\_\allowbreak left", "reward": 0.0, "previous\_\allowbreak transition\_\allowbreak count": 1, "after\_\allowbreak previous\_\allowbreak visits": 8\}}\\[3pt]

\textbf{State novelty} & {\footnotesize\ttfamily \{"reached": \{"compass": "north", "row5": ["unseen", "unseen", "wall", "wall", "wall", "wall", "wall"]\}, "candidate\_\allowbreak previous\_\allowbreak visits": 8, "recent\_\allowbreak state\_\allowbreak example": \{"state": \{"compass": "west", "row5": ["wall", "wall", "wall", "wall", "wall", "unseen", "unseen"]\}, "lifetime\_\allowbreak visits": 1\}\}}\\[3pt]

\textbf{Replay input (shared)} & {\footnotesize\ttfamily \{"before": \{"ahead": "wall", "carrying": "nothing", "visible": []\}, "after": \{"ahead": "wall", "carrying": "nothing", "visible": []\}, "action": "pick\_\allowbreak up", "reward": 0.0, "matching\_\allowbreak transitions": 0, "goal\_\allowbreak entry": false\}}\\[3pt]

\end{tabular}
\end{inputexample}

\clearpage

\begin{inputexample}{SimpleCrossing}

\textbf{Task information.} Find openings through the barriers and reach the green goal.

\textbf{Action choices.} The seven MiniGrid actions above.

\textbf{Concrete scene (action-query example).} {\footnotesize\ttfamily \{"ahead": "empty", "left": "wall", "right": "wall", "carrying": "nothing", "visible": []\}}\par\smallskip
\begin{minipage}[t]{.25\linewidth}\footnotesize\ttfamily
\setlength{\tabcolsep}{3pt}\begin{tabular}{@{}rrrrrrr@{}}
? & ? & ? & ? & ? & ? & ?\\
? & ? & ? & ? & ? & ? & ?\\
? & ? & ? & ? & ? & ? & ?\\
W & W & W & W & W & ? & ?\\
. & . & . & . & W & ? & ?\\
? & ? & W & . & W & ? & ?\\
? & ? & W & . & W & ? & ?\\
\end{tabular}\end{minipage}\hfill\begin{minipage}[t]{.71\linewidth}\footnotesize
The complete native $7\times7$ view for this example. Rows run top to bottom; the agent occupies row 6, column 3 (zero-based), facing row 5. The agent cell encodes its carried object. \texttt{?}: unseen; \texttt{.}: empty; \texttt{W}: wall; \texttt{K}: key; \texttt{D}: door; \texttt{G}: goal; \texttt{L}: lava. Colors and door status are retained in the actual request.
\end{minipage}\par\smallskip

\end{inputexample}
\begin{inputexample}{SimpleCrossing --- method-specific state examples}

\setlength{\tabcolsep}{3pt}\renewcommand{\arraystretch}{1.05}
\begin{tabular}{@{}p{.235\linewidth}p{.735\linewidth}@{}}

\textbf{Action reference} & {\footnotesize\ttfamily \{"current": \{"ahead": "empty", "carrying": "nothing", "visible": []\}\}}\\[3pt]

\textbf{Gate} & {\footnotesize\ttfamily \{"current": \{"ahead": "wall", "carrying": "nothing", "visible": []\}, "greedy\_\allowbreak action": "done", "learner\_\allowbreak updates": 0, "current\_\allowbreak situation\_\allowbreak visits": 7\}}\\[3pt]

\textbf{Action novelty} & {\footnotesize\ttfamily \{"current": \{"ahead": "wall", "carrying": "nothing", "visible": []\}, "action\_\allowbreak to\_\allowbreak rate": "turn\_\allowbreak left", "attempts\_\allowbreak in\_\allowbreak current\_\allowbreak situation": 1, "current\_\allowbreak situation\_\allowbreak visits": 7\}}\\[3pt]

\textbf{Transition diversity} & {\footnotesize\ttfamily \{"before": \{"ahead": "wall", "carrying": "nothing", "visible": []\}, "after": \{"ahead": "wall", "carrying": "nothing", "visible": []\}, "action": "turn\_\allowbreak left", "reward": 0.0, "previous\_\allowbreak transition\_\allowbreak count": 1, "after\_\allowbreak previous\_\allowbreak visits": 8\}}\\[3pt]

\textbf{Replay input (shared)} & {\footnotesize\ttfamily \{"before": \{"ahead": "wall", "carrying": "nothing", "visible": []\}, "after": \{"ahead": "wall", "carrying": "nothing", "visible": []\}, "action": "pick\_\allowbreak up", "reward": 0.0, "matching\_\allowbreak transitions": 0, "goal\_\allowbreak entry": false\}}\\[3pt]

\end{tabular}
\end{inputexample}

\clearpage

\begin{inputexample}{MultiRoom-N2-S4}

\textbf{Task information.} Open encountered doors and navigate through rooms to the goal.

\textbf{Action choices.} The seven MiniGrid actions above.

\textbf{Concrete scene (action-query example).} {\footnotesize\ttfamily \{"ahead": "wall", "left": "wall", "right": "empty", "carrying": "nothing", "visible": []\}}\par\smallskip
\begin{minipage}[t]{.25\linewidth}\footnotesize\ttfamily
\setlength{\tabcolsep}{3pt}\begin{tabular}{@{}rrrrrrr@{}}
? & ? & ? & ? & ? & ? & ?\\
? & ? & ? & ? & ? & ? & ?\\
? & ? & ? & ? & ? & ? & ?\\
? & ? & ? & ? & ? & ? & ?\\
? & ? & ? & ? & ? & ? & ?\\
? & ? & W & W & W & W & ?\\
? & ? & W & . & . & W & ?\\
\end{tabular}\end{minipage}\hfill\begin{minipage}[t]{.71\linewidth}\footnotesize
The complete native $7\times7$ view for this example. Rows run top to bottom; the agent occupies row 6, column 3 (zero-based), facing row 5. The agent cell encodes its carried object. \texttt{?}: unseen; \texttt{.}: empty; \texttt{W}: wall; \texttt{K}: key; \texttt{D}: door; \texttt{G}: goal; \texttt{L}: lava. Colors and door status are retained in the actual request.
\end{minipage}\par\smallskip

\end{inputexample}
\begin{inputexample}{MultiRoom-N2-S4 --- method-specific state examples}

\setlength{\tabcolsep}{3pt}\renewcommand{\arraystretch}{1.05}
\begin{tabular}{@{}p{.235\linewidth}p{.735\linewidth}@{}}

\textbf{Jev-BC / Jev-PEX} & {\footnotesize\ttfamily \{"current": \{"ahead": "wall", "carrying": "nothing", "visible": []\}\}}\\[3pt]

\textbf{Jev action / probability} & {\footnotesize\ttfamily \{"current": \{"ahead": "wall", "carrying": "nothing", "visible": [["red\_\allowbreak closed\_\allowbreak door", 0, 1]]\}\}}\\[3pt]

\textbf{Gate} & {\footnotesize\ttfamily \{"current": \{"ahead": "empty", "carrying": "nothing", "visible": []\}, "greedy\_\allowbreak action": "done", "learner\_\allowbreak updates": 0, "current\_\allowbreak situation\_\allowbreak visits": 2\}}\\[3pt]

\textbf{Action novelty} & {\footnotesize\ttfamily \{"current": \{"ahead": "empty", "carrying": "nothing", "visible": []\}, "action\_\allowbreak to\_\allowbreak rate": "turn\_\allowbreak left", "attempts\_\allowbreak in\_\allowbreak current\_\allowbreak situation": 0, "current\_\allowbreak situation\_\allowbreak visits": 2\}}\\[3pt]

\textbf{Transition diversity} & {\footnotesize\ttfamily \{"before": \{"ahead": "empty", "carrying": "nothing", "visible": []\}, "after": \{"ahead": "wall", "carrying": "nothing", "visible": [["blue\_\allowbreak closed\_\allowbreak door", 0, -1]]\}, "action": "turn\_\allowbreak left", "reward": 0.0, "previous\_\allowbreak transition\_\allowbreak count": 0, "after\_\allowbreak previous\_\allowbreak visits": 7\}}\\[3pt]

\textbf{State novelty} & {\footnotesize\ttfamily \{"reached": \{"compass": "west", "row5": ["unseen", "unseen", "blue\_\allowbreak closed\_\allowbreak door", "empty", "empty", "wall", "unseen"]\}, "candidate\_\allowbreak previous\_\allowbreak visits": 8, "recent\_\allowbreak state\_\allowbreak example": \{"state": \{"compass": "south", "row5": ["unseen", "unseen", "wall", "wall", "blue\_\allowbreak closed\_\allowbreak door", "wall", "unseen"]\}, "lifetime\_\allowbreak visits": 1\}\}}\\[3pt]

\textbf{Replay input (shared)} & {\footnotesize\ttfamily \{"before": \{"ahead": "wall", "carrying": "nothing", "visible": [["blue\_\allowbreak closed\_\allowbreak door", 0, -1]]\}, "after": \{"ahead": "wall", "carrying": "nothing", "visible": [["blue\_\allowbreak closed\_\allowbreak door", 0, -1]]\}, "action": "pick\_\allowbreak up", "reward": 0.0, "matching\_\allowbreak transitions": 0, "goal\_\allowbreak entry": false\}}\\[3pt]

\end{tabular}
\end{inputexample}

\clearpage

\begin{inputexample}{MultiRoom-N4-S5}

\textbf{Task information.} Navigate through rooms; observed-map memory and candidate paths are supplied.

\textbf{Action choices.} The seven MiniGrid actions above.

\textbf{Concrete scene (action-query example).} {\footnotesize\ttfamily \{"ahead": "purple\_\allowbreak open\_\allowbreak door", "left": "empty", "right": "wall", "carrying": "nothing", "visible": [["blue\_\allowbreak closed\_\allowbreak door", 5, -1], ["purple\_\allowbreak open\_\allowbreak door", 1, 0]]\}}\par\smallskip
\begin{minipage}[t]{.25\linewidth}\footnotesize\ttfamily
\setlength{\tabcolsep}{3pt}\begin{tabular}{@{}rrrrrrr@{}}
? & ? & ? & ? & ? & ? & ?\\
? & W & D & W & W & W & ?\\
? & W & . & . & . & W & ?\\
? & W & . & . & . & W & ?\\
? & W & . & . & . & W & ?\\
W & W & W & D & W & ? & ?\\
W & . & . & . & W & ? & ?\\
\end{tabular}\end{minipage}\hfill\begin{minipage}[t]{.71\linewidth}\footnotesize
The complete native $7\times7$ view for this example. Rows run top to bottom; the agent occupies row 6, column 3 (zero-based), facing row 5. The agent cell encodes its carried object. \texttt{?}: unseen; \texttt{.}: empty; \texttt{W}: wall; \texttt{K}: key; \texttt{D}: door; \texttt{G}: goal; \texttt{L}: lava. Colors and door status are retained in the actual request.
\end{minipage}\par\smallskip
\textbf{Observed-map candidate.} {\footnotesize\ttfamily \{"kind": "door", "object": "grey\_\allowbreak open\_\allowbreak door", "cells\_\allowbreak forward": -3, "cells\_\allowbreak right": -1, "currently\_\allowbreak visible": false, "minimum\_\allowbreak primitive\_\allowbreak actions\_\allowbreak on\_\allowbreak observed\_\allowbreak map": 6, "relative\_\allowbreak map\_\allowbreak xy": [0, -2], "times\_\allowbreak entered": 1, "id": "candidate\_\allowbreak 0"\}}\par\smallskip

\end{inputexample}
\begin{inputexample}{MultiRoom-N4-S5 --- method-specific state examples}

\setlength{\tabcolsep}{3pt}\renewcommand{\arraystretch}{1.05}
\begin{tabular}{@{}p{.235\linewidth}p{.735\linewidth}@{}}

\textbf{Jev-BC / Jev-PEX} & {\footnotesize\ttfamily \{"current": \{"ahead": "purple\_\allowbreak open\_\allowbreak door", "carrying": "nothing", "visible": [["blue\_\allowbreak closed\_\allowbreak door", 5, -1], ["purple\_\allowbreak open\_\allowbreak door", 1, 0]]\}\}}\\[3pt]

\textbf{Jev action / probability} & {\footnotesize\ttfamily \{"current": \{"ahead": "wall", "carrying": "nothing", "visible": [["blue\_\allowbreak closed\_\allowbreak door", 0, 3]]\}\}}\\[3pt]

\textbf{Gate} & {\footnotesize\ttfamily \{"current": \{"ahead": "empty", "carrying": "nothing", "visible": [["yellow\_\allowbreak closed\_\allowbreak door", 2, 0]]\}, "greedy\_\allowbreak action": "done", "learner\_\allowbreak updates": 0, "current\_\allowbreak situation\_\allowbreak visits": 2\}}\\[3pt]

\textbf{Action novelty} & {\footnotesize\ttfamily \{"current": \{"ahead": "empty", "carrying": "nothing", "visible": [["yellow\_\allowbreak closed\_\allowbreak door", 2, 0]]\}, "action\_\allowbreak to\_\allowbreak rate": "turn\_\allowbreak left", "attempts\_\allowbreak in\_\allowbreak current\_\allowbreak situation": 0, "current\_\allowbreak situation\_\allowbreak visits": 2\}}\\[3pt]

\textbf{Transition diversity} & {\footnotesize\ttfamily \{"before": \{"ahead": "empty", "carrying": "nothing", "visible": [["yellow\_\allowbreak closed\_\allowbreak door", 2, 0]]\}, "after": \{"ahead": "wall", "carrying": "nothing", "visible": [["yellow\_\allowbreak closed\_\allowbreak door", 0, 2]]\}, "action": "turn\_\allowbreak left", "reward": 0.0, "previous\_\allowbreak transition\_\allowbreak count": 0, "after\_\allowbreak previous\_\allowbreak visits": 6\}}\\[3pt]

\textbf{Replay input (shared)} & {\footnotesize\ttfamily \{"before": \{"ahead": "wall", "carrying": "nothing", "visible": [["yellow\_\allowbreak closed\_\allowbreak door", 0, 2]]\}, "after": \{"ahead": "wall", "carrying": "nothing", "visible": [["yellow\_\allowbreak closed\_\allowbreak door", 0, 2]]\}, "action": "pick\_\allowbreak up", "reward": 0.0, "matching\_\allowbreak transitions": 0, "goal\_\allowbreak entry": false\}}\\[3pt]

\textbf{Navigation: full} & {\footnotesize\ttfamily \{"observed\_\allowbreak memory": "retained", "navigation\_\allowbreak candidates": "supplied", "path\_\allowbreak distance": "supplied"\}}\\[3pt]

\textbf{Navigation: removed} & {\footnotesize\ttfamily \{"observed\_\allowbreak memory": "retained", "navigation\_\allowbreak candidates": "removed", "path\_\allowbreak distance": "removed"\}}\\[3pt]

\end{tabular}
\end{inputexample}

\clearpage

\begin{inputexample}{FourRooms}

\textbf{Task information.} Navigate through wall gaps, not doors; observed-map memory and candidate paths are supplied.

\textbf{Action choices.} The seven MiniGrid actions above.

\textbf{Concrete scene (action-query example).} {\footnotesize\ttfamily \{"ahead": "empty", "left": "empty", "right": "empty", "carrying": "nothing", "visible": []\}}\par\smallskip
\begin{minipage}[t]{.25\linewidth}\footnotesize\ttfamily
\setlength{\tabcolsep}{3pt}\begin{tabular}{@{}rrrrrrr@{}}
. & . & W & . & . & . & .\\
. & . & W & . & . & . & .\\
. & . & W & . & . & . & .\\
. & W & W & W & W & . & W\\
. & . & W & . & . & . & .\\
. & . & W & . & . & . & .\\
. & . & . & . & . & . & .\\
\end{tabular}\end{minipage}\hfill\begin{minipage}[t]{.71\linewidth}\footnotesize
The complete native $7\times7$ view for this example. Rows run top to bottom; the agent occupies row 6, column 3 (zero-based), facing row 5. The agent cell encodes its carried object. \texttt{?}: unseen; \texttt{.}: empty; \texttt{W}: wall; \texttt{K}: key; \texttt{D}: door; \texttt{G}: goal; \texttt{L}: lava. Colors and door status are retained in the actual request.
\end{minipage}\par\smallskip
\textbf{Observed-map candidate.} {\footnotesize\ttfamily \{"kind": "observed\_\allowbreak narrow\_\allowbreak passage", "object": "empty", "cells\_\allowbreak forward": 0, "cells\_\allowbreak right": -1, "currently\_\allowbreak visible": true, "minimum\_\allowbreak primitive\_\allowbreak actions\_\allowbreak on\_\allowbreak observed\_\allowbreak map": 2, "relative\_\allowbreak map\_\allowbreak xy": [3, -6], "times\_\allowbreak entered": 0, "id": "candidate\_\allowbreak 0"\}}\par\smallskip

\end{inputexample}
\begin{inputexample}{FourRooms --- method-specific state examples}

\setlength{\tabcolsep}{3pt}\renewcommand{\arraystretch}{1.05}
\begin{tabular}{@{}p{.235\linewidth}p{.735\linewidth}@{}}

\textbf{Action reference} & {\footnotesize\ttfamily \{"current": \{"ahead": "empty", "carrying": "nothing", "visible": []\}\}}\\[3pt]

\textbf{Gate} & {\footnotesize\ttfamily \{"current": \{"ahead": "empty", "carrying": "nothing", "visible": []\}, "greedy\_\allowbreak action": "done", "learner\_\allowbreak updates": 0, "current\_\allowbreak situation\_\allowbreak visits": 2\}}\\[3pt]

\textbf{Action novelty} & {\footnotesize\ttfamily \{"current": \{"ahead": "empty", "carrying": "nothing", "visible": []\}, "action\_\allowbreak to\_\allowbreak rate": "turn\_\allowbreak left", "attempts\_\allowbreak in\_\allowbreak current\_\allowbreak situation": 0, "current\_\allowbreak situation\_\allowbreak visits": 1\}}\\[3pt]

\textbf{Transition diversity} & {\footnotesize\ttfamily \{"before": \{"ahead": "empty", "carrying": "nothing", "visible": []\}, "after": \{"ahead": "empty", "carrying": "nothing", "visible": []\}, "action": "turn\_\allowbreak right", "reward": 0.0, "previous\_\allowbreak transition\_\allowbreak count": 0, "after\_\allowbreak previous\_\allowbreak visits": 8\}}\\[3pt]

\textbf{State novelty} & {\footnotesize\ttfamily \{"reached": \{"compass": "south", "row5": ["empty", "empty", "empty", "empty", "empty", "empty", "wall"]\}, "candidate\_\allowbreak previous\_\allowbreak visits": 8, "recent\_\allowbreak state\_\allowbreak example": \{"state": \{"compass": "east", "row5": ["empty", "empty", "empty", "empty", "empty", "empty", "wall"]\}, "lifetime\_\allowbreak visits": 1\}\}}\\[3pt]

\textbf{Replay input (shared)} & {\footnotesize\ttfamily \{"before": \{"ahead": "empty", "carrying": "nothing", "visible": []\}, "after": \{"ahead": "empty", "carrying": "nothing", "visible": []\}, "action": "pick\_\allowbreak up", "reward": 0.0, "matching\_\allowbreak transitions": 0, "goal\_\allowbreak entry": false\}}\\[3pt]

\textbf{Subgoal} & {\footnotesize\ttfamily \{"current": \{"ahead": "empty", "left": "empty", "right": "empty", "carrying": "nothing", "visible": []\}, "kind": "visit\_\allowbreak frontier", "object": "empty", "reward\_\allowbreak eligible": true, "completed\_\allowbreak before": false\}}\\[3pt]

\textbf{Navigation: full} & {\footnotesize\ttfamily \{"observed\_\allowbreak memory": "retained", "navigation\_\allowbreak candidates": "supplied", "path\_\allowbreak distance": "supplied"\}}\\[3pt]

\textbf{Navigation: removed} & {\footnotesize\ttfamily \{"observed\_\allowbreak memory": "retained", "navigation\_\allowbreak candidates": "removed", "path\_\allowbreak distance": "removed"\}}\\[3pt]

\end{tabular}
\end{inputexample}

\clearpage

\begin{inputexample}{Seaquest}

\textbf{Task information.} Manage oxygen, rescue divers and avoid enemies; surfacing replenishes oxygen.

\textbf{Action choices.} NOOP, FIRE, UP, RIGHT, LEFT, DOWN, UPRIGHT, UPLEFT, DOWNRIGHT, DOWNLEFT, UPFIRE, RIGHTFIRE, LEFTFIRE, DOWNFIRE, UPRIGHTFIRE, UPLEFTFIRE, DOWNRIGHTFIRE, DOWNLEFTFIRE.

\textbf{Concrete scene (action-query example).} {\footnotesize\ttfamily \{"player\_\allowbreak center": [84.0, 49.5], "object": "Player", "bbox": [76, 46, 16, 7], "oxygen\_\allowbreak fraction": 0.746, "diver\_\allowbreak icons": 0\}}\par\smallskip
{\footnotesize\ttfamily \{"category": "Player", "bbox": [76, 46, 16, 7], "estimated\_\allowbreak displacement\_\allowbreak previous\_\allowbreak four\_\allowbreak frames": [0.0, 0.0]\}}\par
{\footnotesize\ttfamily \{"category": "OxygenBar", "bbox": [49, 170, 47, 5], "estimated\_\allowbreak displacement\_\allowbreak previous\_\allowbreak four\_\allowbreak frames": [1.0, 0.0]\}}\par
\smallskip RGB detections include object categories, boxes and estimated motion; this is a scene excerpt, not an exhaustive detection list. A missing detection does not establish absence.\par\smallskip

\end{inputexample}
\begin{inputexample}{Seaquest --- method-specific state examples}

\setlength{\tabcolsep}{3pt}\renewcommand{\arraystretch}{1.05}
\begin{tabular}{@{}p{.235\linewidth}p{.735\linewidth}@{}}

\textbf{Action reference} & {\footnotesize\ttfamily \{"current": \{"player\_\allowbreak center": [84.0, 49.5], "oxygen\_\allowbreak fraction": 0.746, "diver\_\allowbreak icons": 0\}\}}\\[3pt]

\textbf{Gate} & {\footnotesize\ttfamily \{"current": \{"player\_\allowbreak center": [84.0, 49.5], "oxygen\_\allowbreak fraction": 0.619, "diver\_\allowbreak icons": 0\}, "greedy\_\allowbreak action": "DOWNFIRE", "learner\_\allowbreak updates": 0, "current\_\allowbreak situation\_\allowbreak visits": 4\}}\\[3pt]

\textbf{Action novelty} & {\footnotesize\ttfamily \{"current": \{"player\_\allowbreak center": [84.0, 49.5], "oxygen\_\allowbreak fraction": 0.619, "diver\_\allowbreak icons": 0\}, "action\_\allowbreak to\_\allowbreak rate": "NOOP", "attempts\_\allowbreak in\_\allowbreak current\_\allowbreak situation": 0, "current\_\allowbreak situation\_\allowbreak visits": 4\}}\\[3pt]

\textbf{Transition diversity} & {\footnotesize\ttfamily \{"before": \{"player\_\allowbreak center": [84.0, 49.5], "oxygen\_\allowbreak fraction": 0.619, "diver\_\allowbreak icons": 0\}, "after": \{"player\_\allowbreak center": [84.0, 49.5], "oxygen\_\allowbreak fraction": 0.651, "diver\_\allowbreak icons": 0\}, "action": "UP", "reward": 0.0, "previous\_\allowbreak transition\_\allowbreak count": 0, "after\_\allowbreak previous\_\allowbreak visits": 4\}}\\[3pt]

\textbf{State novelty} & {\footnotesize\ttfamily \{"reached": \{"coarse\_\allowbreak situation": \{"game": "Seaquest", "player\_\allowbreak region": [5, 3], "oxygen\_\allowbreak band": 3, "diver\_\allowbreak icons": 0, "surface": true\}\}, "candidate\_\allowbreak previous\_\allowbreak visits": 0, "recent\_\allowbreak state\_\allowbreak example": \{"state": \{"diver\_\allowbreak icons": 0, "game": "Seaquest", "oxygen\_\allowbreak band": 2, "player\_\allowbreak region": [5, 3], "surface": true\}, "lifetime\_\allowbreak visits": 8\}\}}\\[3pt]

\textbf{Replay input (shared)} & {\footnotesize\ttfamily \{"before": \{"player\_\allowbreak center": [84.0, 49.5], "oxygen\_\allowbreak fraction": 0.397, "diver\_\allowbreak icons": 0\}, "after": \{"player\_\allowbreak center": [84.0, 49.5], "oxygen\_\allowbreak fraction": 0.429, "diver\_\allowbreak icons": 0\}, "action": "UPFIRE", "reward": 0.0, "matching\_\allowbreak transitions": 0, "goal\_\allowbreak entry": false\}}\\[3pt]

\textbf{Subgoal} & {\footnotesize\ttfamily \{"current": \{"player\_\allowbreak center": [23.0, 90.0], "object": "Player", "bbox": [9, 85, 28, 10], "oxygen\_\allowbreak fraction": 0.937, "diver\_\allowbreak icons": 1\}, "candidate\_\allowbreak goals": ["dive\_\allowbreak deep", "carry\_\allowbreak two", "surface\_\allowbreak with\_\allowbreak diver"], "recent\_\allowbreak attempts": [\{"goal": "dive\_\allowbreak mid", "success": true, "steps": 44\}]\}}\\[3pt]

\end{tabular}
\end{inputexample}

\clearpage

\begin{inputexample}{Freeway}

\textbf{Task information.} Move the left chicken to the top; cars push it backward on collision.

\textbf{Action choices.} NOOP, UP, DOWN.

\textbf{Concrete scene (action-query example).} {\footnotesize\ttfamily \{"object": "player\_\allowbreak chicken", "bbox": [44, 155, 6, 8]\}}\par\smallskip
{\footnotesize\ttfamily \{"category": "player\_\allowbreak chicken", "bbox": [44, 155, 6, 8], "estimated\_\allowbreak displacement\_\allowbreak previous\_\allowbreak four\_\allowbreak frames": [-0.5, -4.0]\}}\par
{\footnotesize\ttfamily \{"category": "car\_\allowbreak lane\_\allowbreak 8", "bbox": [19, 156, 8, 8], "estimated\_\allowbreak displacement\_\allowbreak previous\_\allowbreak four\_\allowbreak frames": [1.0, 0.0]\}}\par
\smallskip RGB detections include object categories, boxes and estimated motion; this is a scene excerpt, not an exhaustive detection list. A missing detection does not establish absence.\par\smallskip

\end{inputexample}
\begin{inputexample}{Freeway --- method-specific state examples}

\setlength{\tabcolsep}{3pt}\renewcommand{\arraystretch}{1.05}
\begin{tabular}{@{}p{.235\linewidth}p{.735\linewidth}@{}}

\textbf{Jev-BC / Jev-PEX} & {\footnotesize\ttfamily \{"current": \{"object": "player\_\allowbreak chicken", "bbox": [44, 155, 6, 8]\}\}}\\[3pt]

\textbf{Jev action / probability} & {\footnotesize\ttfamily \{"current": \{"player\_\allowbreak center": [47.0, 133.5]\}\}}\\[3pt]

\textbf{Gate} & {\footnotesize\ttfamily \{"current": \{"player\_\allowbreak center": [47.0, 183.0]\}, "greedy\_\allowbreak action": "UP", "learner\_\allowbreak updates": 0, "current\_\allowbreak situation\_\allowbreak visits": 17\}}\\[3pt]

\textbf{Action novelty} & {\footnotesize\ttfamily \{"current": \{"player\_\allowbreak center": [47.0, 183.0]\}, "action\_\allowbreak to\_\allowbreak rate": "NOOP", "attempts\_\allowbreak in\_\allowbreak current\_\allowbreak situation": 3, "current\_\allowbreak situation\_\allowbreak visits": 17\}}\\[3pt]

\textbf{Transition diversity} & {\footnotesize\ttfamily \{"before": \{"player\_\allowbreak center": [47.0, 183.0]\}, "after": \{"player\_\allowbreak center": [47.0, 183.0]\}, "action": "NOOP", "reward": 0.0, "previous\_\allowbreak transition\_\allowbreak count": 3, "after\_\allowbreak previous\_\allowbreak visits": 17\}}\\[3pt]

\textbf{State novelty} & {\footnotesize\ttfamily \{"reached": \{"coarse\_\allowbreak situation": \{"game": "Freeway", "player\_\allowbreak region": [2, 11]\}\}, "candidate\_\allowbreak previous\_\allowbreak visits": 17, "recent\_\allowbreak state\_\allowbreak example": \{"state": \{"game": "Freeway", "player\_\allowbreak region": [2, 11]\}, "lifetime\_\allowbreak visits": 17\}\}}\\[3pt]

\textbf{Replay input (shared)} & {\footnotesize\ttfamily \{"before": \{"player\_\allowbreak center": [47.0, 183.0]\}, "after": \{"player\_\allowbreak center": [47.5, 179.0]\}, "action": "UP", "reward": 0.0, "matching\_\allowbreak transitions": 4, "goal\_\allowbreak entry": false\}}\\[3pt]

\textbf{Subgoal} & {\footnotesize\ttfamily \{"current": \{"player\_\allowbreak center": [47.0, 191.0], "object": "player\_\allowbreak chicken", "bbox": [44, 187, 6, 8]\}, "candidate\_\allowbreak goals": ["reach\_\allowbreak y116", "reach\_\allowbreak y68", "cross"], "recent\_\allowbreak attempts": [\{"goal": "reach\_\allowbreak y164", "success": true, "steps": 9\}]\}}\\[3pt]

\end{tabular}
\end{inputexample}

\clearpage

\begin{inputexample}{SpaceInvaders}

\textbf{Task information.} Move the cannon, shoot aliens and avoid incoming projectiles.

\textbf{Action choices.} NOOP, FIRE, RIGHT, LEFT, RIGHTFIRE, LEFTFIRE.

\textbf{Concrete scene (action-query example).} {\footnotesize\ttfamily \{"object": "player", "bbox": [34, 185, 7, 10]\}}\par\smallskip
{\footnotesize\ttfamily \{"category": "player", "bbox": [34, 185, 7, 10]\}}\par
{\footnotesize\ttfamily \{"category": "alien", "bbox": [23, 31, 8, 10], "observed\_\allowbreak displacement\_\allowbreak previous\_\allowbreak four\_\allowbreak frames": [1.0, 0.0]\}}\par
{\footnotesize\ttfamily \{"category": "shield", "bbox": [42, 157, 8, 18], "observed\_\allowbreak displacement\_\allowbreak previous\_\allowbreak four\_\allowbreak frames": [0.0, 0.0]\}}\par
\smallskip RGB detections include object categories, boxes and estimated motion; this is a scene excerpt, not an exhaustive detection list. A missing detection does not establish absence.\par\smallskip

\end{inputexample}
\begin{inputexample}{SpaceInvaders --- method-specific state examples}

\setlength{\tabcolsep}{3pt}\renewcommand{\arraystretch}{1.05}
\begin{tabular}{@{}p{.235\linewidth}p{.735\linewidth}@{}}

\textbf{Action reference} & {\footnotesize\ttfamily \{"current": \{"object": "player", "bbox": [34, 185, 7, 10]\}\}}\\[3pt]

\textbf{Gate} & {\footnotesize\ttfamily \{"current": \{"object": "alien", "bbox": [22, 31, 8, 10]\}, "greedy\_\allowbreak action": "LEFT", "learner\_\allowbreak updates": 0, "current\_\allowbreak situation\_\allowbreak visits": 9\}}\\[3pt]

\textbf{Action novelty} & {\footnotesize\ttfamily \{"current": \{"object": "alien", "bbox": [22, 31, 8, 10]\}, "action\_\allowbreak to\_\allowbreak rate": "NOOP", "attempts\_\allowbreak in\_\allowbreak current\_\allowbreak situation": 1, "current\_\allowbreak situation\_\allowbreak visits": 9\}}\\[3pt]

\textbf{Transition diversity} & {\footnotesize\ttfamily \{"before": \{"object": "alien", "bbox": [22, 31, 8, 10]\}, "after": \{"object": "alien", "bbox": [22, 31, 8, 10]\}, "action": "NOOP", "reward": 0.0, "previous\_\allowbreak transition\_\allowbreak count": 0, "after\_\allowbreak previous\_\allowbreak visits": 9\}}\\[3pt]

\textbf{State novelty} & {\footnotesize\ttfamily \{"reached": \{"coarse\_\allowbreak situation": \{"game": "SpaceInvaders", "player\_\allowbreak region": [2, 11]\}\}, "candidate\_\allowbreak previous\_\allowbreak visits": 8, "recent\_\allowbreak state\_\allowbreak example": \{"state": \{"game": "SpaceInvaders", "player\_\allowbreak region": "unknown"\}, "lifetime\_\allowbreak visits": 9\}\}}\\[3pt]

\textbf{Replay input (shared)} & {\footnotesize\ttfamily \{"before": \{"object": "alien", "bbox": [22, 31, 8, 10]\}, "after": \{"player\_\allowbreak center": [37.5, 190.0]\}, "action": "LEFT", "reward": 0.0, "matching\_\allowbreak transitions": 0, "goal\_\allowbreak entry": false\}}\\[3pt]

\end{tabular}
\end{inputexample}

\clearpage
\begin{inputexample}[colback=orange!2!white,colbacktitle=orange!10!white,colframe=orange!50!black]{Montezuma's Revenge: image PPO+RND subgoals}
The query supplies RGB-derived regions, visible key/inventory evidence,
first-room geometry and explicit death rules. It asks for a milestone; the
image-only PPO+RND controller chooses primitive actions.

\textbf{State example.} {\footnotesize\ttfamily \{"player\_\allowbreak x\_\allowbreak interval": [80, 95], "player\_\allowbreak feet\_\allowbreak y\_\allowbreak interval": [128, 143], "floor\_\allowbreak key\_\allowbreak detected": true, "key\_\allowbreak inventory": "unknown\_\allowbreak or\_\allowbreak not\_\allowbreak confirmed", "skull\_\allowbreak region": "center"\}}

\textbf{Candidate example.} {\footnotesize\ttfamily \{"collect\_\allowbreak key": "Obtain a visually confirmed key HUD icon; proximity alone is insufficient.", "central\_\allowbreak middle": "Reach landmark x=80, feet\_\allowbreak y=136 with stable position and survive."\}}

\textbf{Other candidate landmarks.} right\_middle, right\_floor, left\_floor, left\_middle, central\_upper, left\_door\_approach, right\_door\_approach.

\textbf{Rule excerpt.} A visible floor key is not carried inventory. A locked door requires the key. Lethal creatures, pits and long unsupported falls can lose a life; use ladders or rope for controlled descent.
\end{inputexample}

\begin{inputexample}{Pure Jev and the matched Qwen interface}
Pure Jev uses the corresponding action description and executes its most likely
action. Qwen receives the same structured state, task question and legal-action
descriptions, serialized as its user message. For example, the response schema
requires \texttt{\{"action": "move\_forward"\}} with the value restricted to
the seven legal MiniGrid labels; this illustrates the format, not a fixed action.

Qwen uses \texttt{model=qwen3-8b}, temperature zero, a 48-token limit and thinking
disabled. Its system instruction is: ``Select one legal action using only the
supplied situation, question, and option descriptions. Return a JSON object with
the single key action. Do not output explanations.'' FourRooms and MultiRoom-N4
use the full-navigation setting above. DoorKey-5/8, Empty, LavaGap, MultiRoom-N2 and
SimpleCrossing use their corresponding direct-Jev local-observation setting.
\end{inputexample}

\end{document}